\documentclass[11pt]{article}
\pdfoutput=1
\usepackage[utf8]{inputenc}
\usepackage[OT1]{fontenc}
\usepackage[usenames]{color}

\usepackage[protrusion=true,expansion=true,final,babel]{microtype}

\usepackage{fullpage}

\usepackage[colorlinks,linkcolor=red,anchorcolor=blue,citecolor=blue]{hyperref}

\usepackage[numbers]{natbib}

\usepackage[subcaption]{styles/smile}
\usepackage{styles/cme-math}
\usepackage{cancel}

\crefname{lemma}{lemma}{lemmas}
\Crefname{lemma}{Lemma}{Lemmas}

\begin{document}

\title{Learning an Inventory Control Policy with General Inventory Arrival Dynamics}

\author{
	Sohrab Andaz\thanks{Amazon, New York, NY. Correspondence to: \href{mailto:sandaz@amazon.com}{sandaz@amazon.com}.}
	\and
	Carson Eisenach\footnotemark[1]
	\and
	Dhruv Madeka\footnotemark[1]
	\and
	Kari Torkkola\footnotemark[1]
	\and
	Randy Jia\thanks{Afresh}~\thanks{Work done while at Amazon.}
	\and
	Dean Foster\footnotemark[1]
	\and
	Sham Kakade\footnotemark[1]~\thanks{Harvard University, Cambridge, MA.}
}

\maketitle

\begin{abstract}
	In this paper we address the problem of learning and backtesting inventory control policies in the presence of general arrival dynamics -- which we term as a quantity-over-time arrivals model (QOT).  We also allow for order quantities to be modified as a post-processing step to meet vendor constraints such as order minimum and batch size constraints -- a common practice in real supply chains. To the best of our knowledge this is the first work to handle either arbitrary arrival dynamics or an arbitrary downstream post-processing of order quantities. Building upon recent work \citep{madeka2022deep} we similarly formulate the periodic review inventory control problem as an {\it exogenous decision process}, where most of the state is outside the control of the agent. \citet{madeka2022deep} show how to construct a simulator that replays historic data to solve this class of problem. In our case, we incorporate a deep generative model for the arrivals process as part of the history replay. By formulating the problem as an exogenous decision process, we can apply results from \citet{madeka2022deep} to obtain a reduction to supervised learning. Via simulation studies we show that this approach yields statistically significant improvements in profitability over production baselines. Using data from a real-world A/B test, we show that Gen-QOT generalizes well to off-policy data and that the resulting buying policy outperforms traditional inventory management systems in real world settings.
\end{abstract}


\section{Introduction}
\label{sec:intro}

The periodic review inventory control problem is that of determining how much inventory to hold in order to maximize revenue. This problem has been studied extensively in the operations research literature, and is often formulated as a Markov decision process \citep{porteus2002foundations}. Some complexities involved include stochastic demands with unknown seasonality, lost sales, stochastic vendor lead times, multiple shipments per order, unreliable replenishment, and order quantity restrictions. Classical approaches are typically able to handle only a subset of these complexities due to the curse of dimensionality and the fact that closed form solutions are not available as the problem setting increases in complexity. In fact, base stock policies (which are optimal for simplified settings and often used in practice) have been shown to perform worse than constant order policies in the presence of lost sales and stochastic demand \citep{zipkin2008old}.

\citet{madeka2022deep} recently introduced the first Deep Reinforcement Learning periodic review inventory system which was able to handle many of the aforementioned challenges. By formulating inventory control as an {\it exogenous decision process} \citep{sinclair2023hindsight}, \citet{madeka2022deep} demonstrate a reduction in complexity of the learning problem to that of supervised learning. They also show how censored (unobserved) historic data can be  used for policy learning and backtesting by constructing a simulator that {\it replays historic data} rather than assuming a parametric form in order to simulate the future.

\begin{figure}[H]
	\centering
	\subcaptionbox{Number of distinct shipments per order\label{fig:multipleshipment-receives}}[0.49\textwidth]
	{\includegraphics[width=0.49\textwidth]{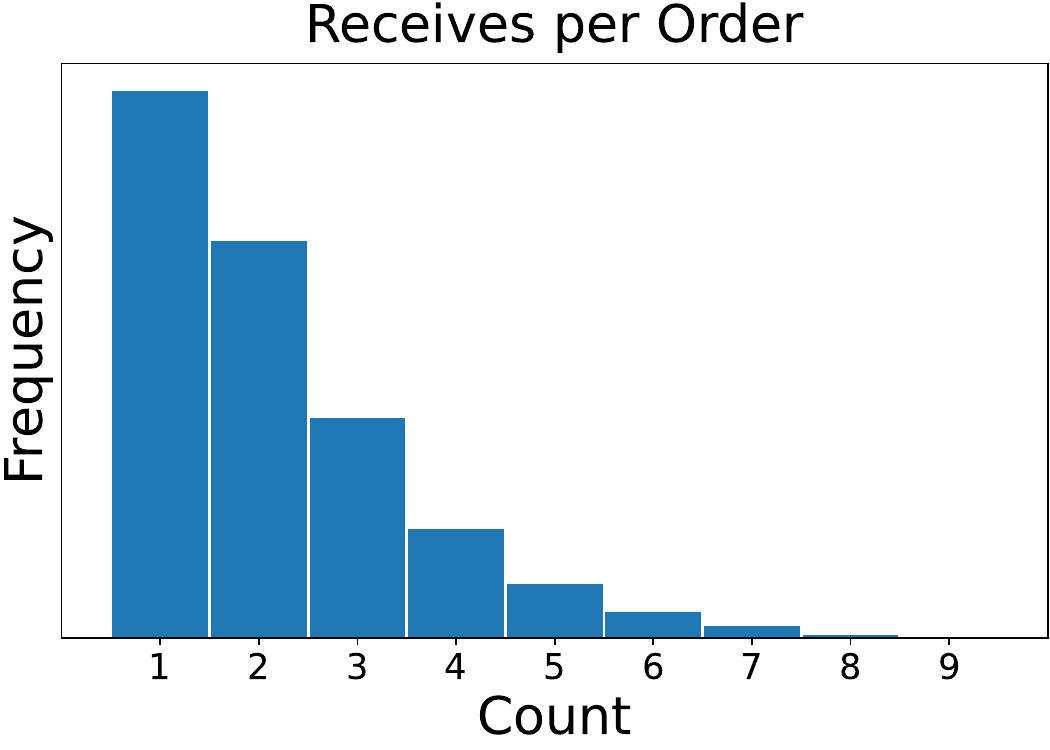}}~
	\subcaptionbox[0.9\textwidth]{Inter-arrival times for the same order \label{fig:multipleshipment-gaps}}
	{\includegraphics[width=0.49\textwidth]{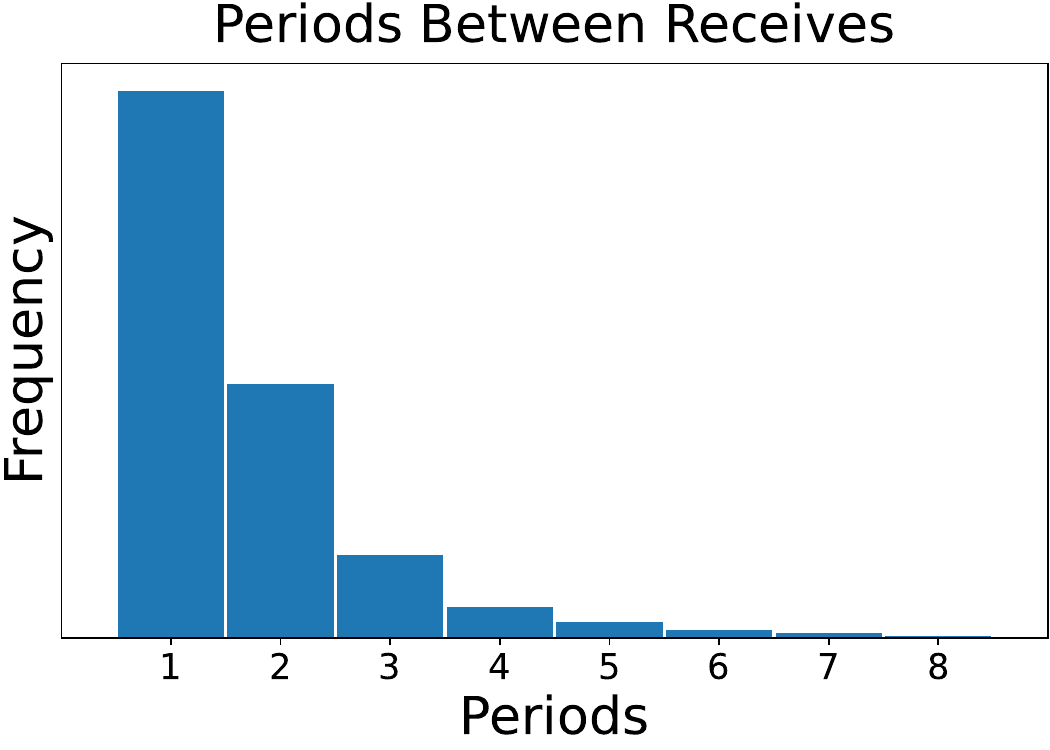}}
	\caption{Orders often arrive in multiple shipments, and spread out over multiple time periods.}
	\label{fig:multipleshipment}
\end{figure}

One of the assumptions \citet{madeka2022deep} does maintain from the classical inventory control literature, however is the structure of inventory arrivals given an order quantity provided to the vendor. The authors handle the case of stochastic lead times (with unknown distribution), but in real-world settings there are several additional complexities that can arise. First, a single order placed to a vendor may arrive in multiple shipments at different future periods\footnote{To the best of our knowledge, no existing work considers this setting.}. \Cref{fig:multipleshipment} shows distributions of number of shipments and inter-arrival times from a set of purchase orders made by a large e-retailer.

\begin{figure}[h]
	\centering
	\subcaptionbox{Observed distribution of supplier fill rates\label{fig:fills-vendor}}[0.49\textwidth]
	{\includegraphics[width=0.49\textwidth]{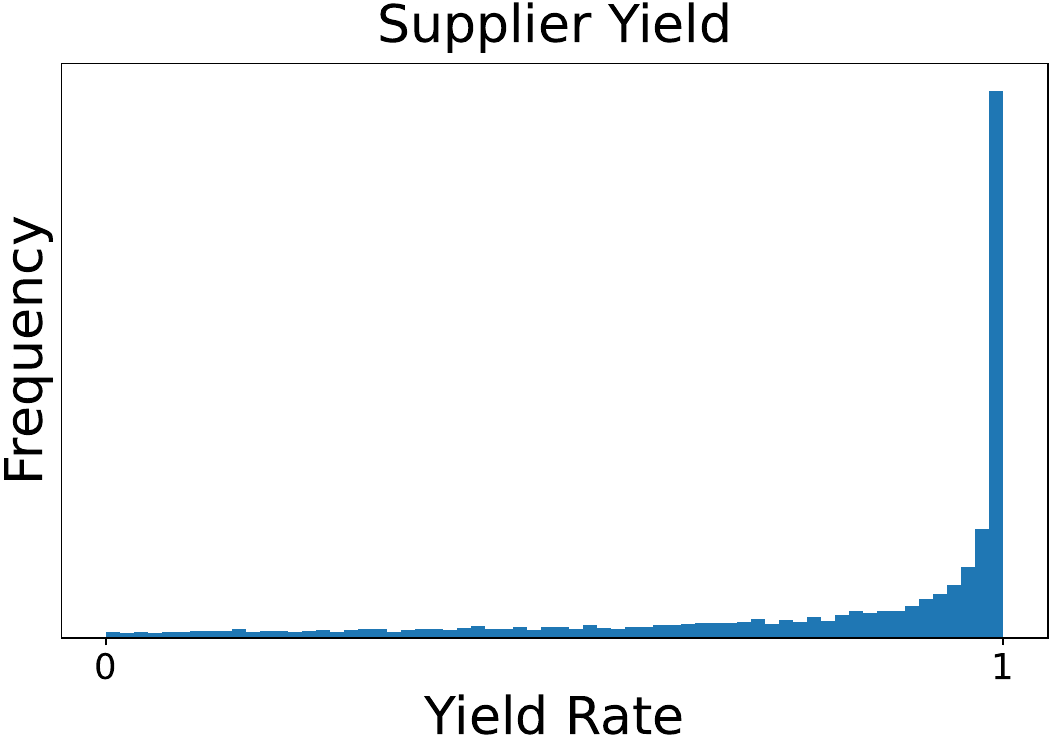}}
	\subcaptionbox[0.49\textwidth]{Overall yields, including any post-processing \label{fig:fills-system}}
	{\includegraphics[width=0.49\textwidth]{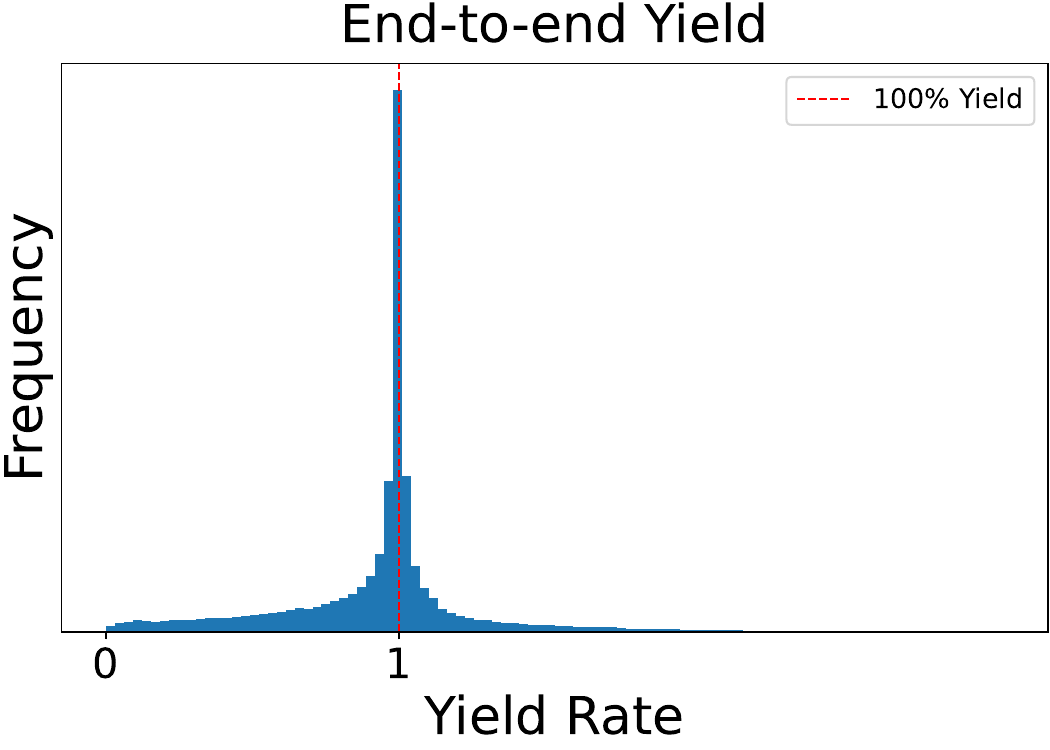}}
	\caption{Example of yields -- multiplicative proportion of order quantity that is received -- for purchase orders at a large e-retailer, both before and after any post-processor is applied to the order quantity.}
	\label{fig:fills}
\end{figure}

Second, the supply may be unreliable and vendors may only partially fill orders that they receive -- for example an order for 100 units of an item may result in only 75 units being supplied. This may occur for multiple reasons, including that the vendor itself is out of stock. In the literature the proportion of the original order quantity retailer ultimately receives is referred to as the {\it yield} or {\it fill rate}. \Cref{fig:fills-vendor} shows the distribution of yields at a large e-retailer.

Finally, the order quantity dictated by the policy may not meet the requirements of the vendor and may need to be rounded before a purchase order is sent: for example minimum order quantities or batch size restrictions\footnote{For example, order quantities may need to be a multiple of case sizes or other unit of packaging} are quite common \citep{zhu2015effective}. The impact of such restrictions is difficult to analyze, is not well understood, and heuristic rules are typically employed \citep{robb1998inventory,zhu2015effective,zhou2007effective}. In real-world settings, this may be performed as a secondary post-processing step after the optimal order quantity has been determined -- for example by rounding up to the minimum order quantity \citep{kiesmller2011single,zhu2022simple}. \Cref{fig:fills-system} shows the distribution of ``end-to-end'' yields -- including both supply uncertainty and any post-processing steps to meet batch ordering or minimum order quantity constraints applied by ordering systems -- for replenishment decisions at a large e-retailer. Note how in many cases, the ``end-to-end'' yield can be greater than the original order quantity due to the presence of a post-processing step.

\subsubsection*{Our Contributions and Organization}

\begin{figure}[h]
	\centering
	\includegraphics[width=0.5\textwidth]{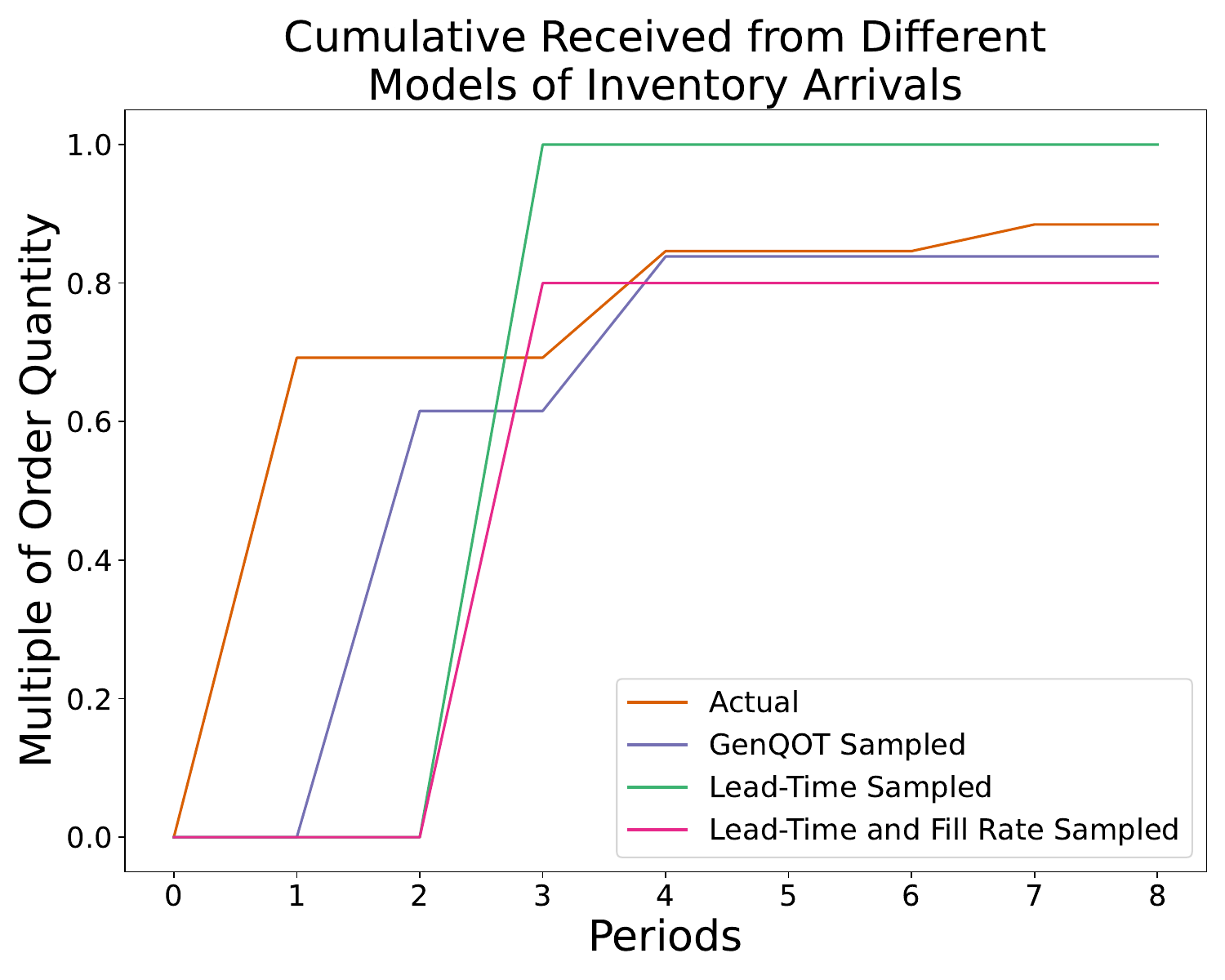}
	\caption{Cumulative arrivals over time for a single purchase order under differing dynamics.}
	\label{fig:diff-models}
\end{figure}

In this paper we address the problem of learning and backtesting inventory control policies in the presence of general arrival dynamics (which we term as a quantity-over-time model or QOT).  We also allow for order quantities to be modified as a post-processing step to meet vendor constraints such as order minimum and batch size constraints. To the best of our knowledge this is the first work to handle either arbitrary arrival dynamics or an arbitrary downstream post-processing of order quantities -- a common practice in real supply chains. \Cref{fig:diff-models} illustrates the difference in cumulative arrivals over time under the different types of arrival dynamics.

The remainder of this paper is organized as follows: in \Cref{sec:method} we formulate our problem as an exogenous interactive decision process and leverage results from \citet{madeka2022deep} to demonstrate a reduction to supervised learning. We also describe our approach to modeling the arrival dynamics for use as part of a simulator that replays historic data \citep{madeka2022deep}. Next, in \Cref{sec:eval-dynamics} we evaluate the performance of the learned dynamics model on  ``on-policy'' data -- that is, data generated by the same policy that generated the data used to fit the model.  Then, \Cref{sec:eval-sim} we demonstrate via backtests the impact a realistic arrivals model has on policy performance. Finally, in \Cref{sec:eval-real}, results from a large real-world A/B test in the supply chain of a large e-retailer show that (1) the RL policy learned using our methodology outperforms classic approaches to periodic review inventory control, and (2) our learned dynamics model generalizes well to ``off-policy'' data which validates our assumption that we have an accurate forecast of transitions under the learned policy.

\section{Related Work}
\label{sec:background}

\subsection{Forecasting and Generative Modeling}
\label{sec:background-fcst}

All work which considers stochastic vendor lead times implicitly requires a forecast of lead times (even if the lead time distribution is assumed to be stationary). We have not found much study of forecasting lead times specifically, but approaches from the probabilistic time series forecasting community can be used to forecast lead times.  There is an extensive body of work which has successfully applied deep learning to time series forecasting, including in the supply chain for forecasting demand   \citep{nascimento2019stconv,yu2017long,gasparin2019deep,mukhoty2019seq,wen2017mqcnn,salinas2020deepar,wen2019deep,eisenach2020mqt,madeka2018sample}.

\subsubsection*{Deep Generative Modeling}
A number of deep generative techniques have been developed to estimate the likelihood of observations in training data and generate new samples from the underlying data distribution. These include generative adversarial networks \citep{goodfellow2014gan}, variational auto-encoders \citep{kingma2013auto,kingma2019intro}, and autoregressive models. Autoregressive models have been used successfully in image generation \citep{van2016pixel}, NLP \citep{brown2020language,devlin2019bert}, and time-series forecasting \citep{madeka2018sample,eisenach2020mqt,januschowski2022forecasting,wen2017mqcnn,salinas2020deepar,lim2019tft}. Our work employs autoregressive modeling by decomposing the full problem of estimating the joint distribution of arrivals into the simpler problem of merely predicting the next arrival in a sequence given the previously realized shipments.

\nocite{augenblick2018belief,taleb2018election,taleb2019all,foster2021threshold}




\subsection{Reinforcement Learning and Exogenous Sequential Decision Problems}
\label{sec:background-exo}
Reinforcement learning has been applied to many sequential decision-making problems including games and simulated physics models where many simulations are possible \citep{silver2016mastering,szepesvari2010algorithms,mnih2013playing, sutton2020reinforcement,schulman2017proximal,mnih2016asynchronous}. In general, one can require exponentially many samples (in the horizon of the problem) to learn a control policy.

Recently, \citet{madeka2022deep} and \citet{sinclair2023hindsight} considered a class of problems called {\it exogenous interactive decision processes} wherein the state consists of a stochastic exogenous process (independent of the control) and an endogenous component that is governed by a known transition function $f$ of both the previous endogenous state and the exogenous process. In these cases, \citet{madeka2022deep,sinclair2023hindsight} prove a reduction in sample complexity to that of supervised learning -- an exponential improvement over the general RL setting. \nocite{efroni2022sample,efroni2022sparsity}

\subsection{Periodic Review Inventory Systems}
\label{sec:background-inv}
Inventory control systems have been studied extensively in the literature under a variety of conditions (see \citet{porteus2002foundations} for a comprehensive overview). The simplest form is the newsvendor, which solves a myopic problem \citep{arrow1958studies}. Many extensions exist \citep{nahmias1979simple, arrow1958studies}, and the optimal policy in many variants takes the form of a {\it base stock policy} which, informally, consists of an optimal inventory level and then orders up to that level. However, outside the restrictive conditions under which optimal policies can be derived, \citet{zipkin2008old} showed that even constant order policies can be better than base stock policies.

\subsubsection*{RL for Inventory Control}
More recently, several authors have applied reinforcement learning to solve multi-period inventory control problems \citep{giannoccaro2002inventory,das1999solving,balaji2019orl,madeka2022deep}. We adopt the modeling approach of \citet{madeka2022deep}, and similarly treat the periodic review inventory control problem under general arrival dynamics as an exogenous decision process. Also following \citet{madeka2022deep}, we build a differentiable simulator \citep{suh2022differentiable,hu2019chainqueen,ingraham2018learning,clavera2020model} using historical supply chain data, and successfully train and backtest an RL agent to achieve improved real world performance over traditional methods.

\nocite{alvo2023neural,mousa2023analysis,abbas2023plasticity,zhao2023policy,thomas2023towards,parmas2023model,gijsbrechts2022can, qi2023practical}

\subsubsection*{Lead Times and Supply Reliability}
There has been extensive work in the literature on how to handle stochastic lead times in periodic review inventory systems \citep{porteus2002foundations,song1996inventory,maggiar2022multi,nahmias1979simple,kaplan1970dynamic}. In addition, unreliable supply has been studied in a variety of settings. One form this often takes is a stochastic yield \citep{maddah2008economic,li2004periodic,henig1990structure, gerchak1988periodic,maggiar2022multi,bollapragada1999myopic}, for example via some portion of the supply being defective. A common heuristic is to simply adjust the order quantity for the mean yield \citep{bollapragada1999myopic}. The second standard formulation is to have the fill rate level be determined by the number of units the vendor has available, where vendors fill up to the amount of units they have available \citep{dada2007news}. The first setting is more common as it is more tractable to find analytic solutions. We are unaware of any work that considers the impact of a varying number of shipments per order placed with the supplier.

In this work, we take the following approach: we assume that the vendor fills orders up to their capacity (which is exogenous to the retailer's replenishment decisions), but we add an arrival shares process which encodes how the filled amount arrives over future time periods.

\subsubsection*{Minimum Order Quantities, Maximum Order Quantities and Batch Ordering}
A {\it minimum order quantity} or {\it maximum order quantity} constraint means that the supplier will reject all orders under or over that amount, respectively. For minimum order quantity constraints, {\it we only consider the setting where the retailer can also choose to order 0 (i.e. not to order)} as otherwise the minimum quantity constraint is unimportant. Similarly, a {\it batch ordering} requirement means that the supplier only accepts order quantities that are an integer multiple of some specified batch quantity. Maximum constraints have been studied extensively in the literature and typically affect the policy a in benign way \citep{chan1999effects,federgruen1986inventory}. On the other hand, minimum order quantities \citep{fisher1996reducing,zhou2007effective,zhao2006structure,shen2019two} and batch requirements \citep{veinott1965optimal,zhu2015effective} -- although  widely adopted by suppliers in practice \citep{zhu2022simple} --  are relatively unstudied in the literature as they present significant difficulties in deriving the optimal order quantity. For minimum order quantities, even in highly simplified settings, optimal policies are only partially characterized and are too complicated to implement in practice \citep{zhao2006structure}. Instead, retailers use base stock policies with a heuristic rounding before placing the order \citep{zhou2007effective,kiesmller2011single}.

\section{Mathematical Formulation and Methodology}
\label{sec:method}

In this section, we follow the Interactive Decision Process (IDP) formulation of \citet{madeka2022deep}, borrowing most of the conventions and notation, except we define and treat the ``lead time'' process differently. At a high level, a central planner is trying to determine how many units of inventory to order at every time step $t = 1,2,...,T$, in order to satisfy demands $D_t$. The goal is to maximize profits by balancing having enough inventory on hand to satisfy demand (we assume a lost sales customer model), with the cost of holding too much inventory.

The dynamics we are most concerned with is how order quantities selected by the policy evolve into future arrivals at the retailer's warehouse. The standard formulation in the literature is the vendor lead time (VLT) arrivals model, whereupon placing an inventory order decision $a_t$ at time $t$, a single quantity $v_t$ is drawn from an exogenous lead time distribution, and the entire order arrives $v_t$ time steps later at time $t+v_t$. In the case of stochastic yields, there are two approaches in the literature: either the yield is a random multiplicative factor multiplied times the order quantity, or the vendor has a stochastic supply and fills up to the amount of their supply. Where we depart from \citet{madeka2022deep} is that our formulation allows that
\begin{enumerate}[itemsep=0em]
	\item inventory can arrive in multiple shipments for a single order,
	\item yields can be stochastic, and
	\item there can be a downstream system that applies a heuristic order quantity rounding to satisfy batch, minimum and maximum quantity constraints.
\end{enumerate}

We propose a novel quantity over time (QOT) arrivals model, which generalizes all the settings described above. In the QOT arrivals model, we assume that orders can arrive in multiple shipments over time, and the total arriving quantity may not necessarily sum up to the order quantity placed. At every time $t$, the vendor has allocated a supply $U_t$ that denotes the maximum number of units it can send (regardless the amount we order), which will arrive over from the current week up to $L$ weeks in the future according to an exogenous arrival shares vector $(\rho_{t,0},...,\rho_{t,L})$ where $\sum_l \rho_{t,l} = 1$. That is, the arrivals at lead time $j$ from order $a_t$ is equal to $\min(U_t, a_t)\rho_{t,j}$. We denote the arrival quantity as $o_{t,j} := \min(U_{t}, a_{t})\rho_{t,j}$. Here, we implicitly assume that orders necessarily arrive after $L$ time steps.

We also allow for the existence of an order quantity post-processor $f_{p}$ that is arbitrary (but known) that modifies order quantities before they are sent to the supplier -- e.g. to ensure they meet any vendor constraints.

\subsection{Mathematical notation}
Denote by $\mathbb{R}$, $\mathbb{R}_{\geq 0}$, $\mathbb{Z}$, and $\mathbb{Z}_{\geq 0}$ the set of reals, non-negative reals, integers, and non-negative integers, respectively. We let $(\cdot)^+$ refer to the classical positive part operator i.e. $(\cdot)^{+} = \max(\cdot, 0)$. 
Let $[\;\cdot\;]$ refer to the set of positive integers up to the argument, i.e. $[\;\cdot\;] = \{x \in \mathbb{Z} \, | \, 1 \leq x \leq \cdot \;\}$. The inventory management problem seeks to find the optimal inventory level for each product $i$ in the set of retailer's products, which we denote by $\mathcal{A}$. We assume our exogenous random variables are defined on a canonical probability space $(\Omega, \mathcal{F}, \mathbb{P})$, and policies are parameterized by $\theta$  in some parameter set $\Theta$.  We use $\mathbb{E}^{\mathbb{P}}$ to denote an expectation operator of a random variable with respect to some probability measure $\mathbb{P}$. Let $||X,Y||_{TV}$ denote the total variation distance between two probability measures $X$ and $Y$.

\subsection{IDP Construction}
\label{sec:idpconstruction}
Our IDP is governed by external (exogenous) processes, a control process, inventory evolution dynamics, and a reward function. To succinctly describe our process, we focus on just one product $i \in \mathcal{A}$, though we note that decisions can be made jointly for every product.

\paragraph{External Processes} At every time step $t$, for product $i$, we assume that there is a random demand process $D_t^{i} \in [0,\infty)$ that corresponds to customer demand during time $t$ for product $i$. We also assume that the random variables $p_t^{i} \in [0,\infty)$ and $c_t^{i} \in [0,\infty)$ are the random variables corresponding to selling price and purchase cost. We also assume any constraints the vendor imposes on the retailer's orders $\bM^i_t \in \RR^{d_v}$ -- such as minimum order quantities and batch sizes -- are exogenous to the ordering decisions. The supply $U^i_t \in [0,\infty)$ corresponds to the maximum amount of inventory the vendor is able to send. Finally, the arrival shares process $\boldsymbol\rho^i_t := \{\rho^i_{t,j}\}_{j=0}^L$ describes the arrivals over the next $L$ time steps from an order placed at the current time $t$ -- note that $\sum_{j=0}^L \rho^i_{t,j} = 1$ and $\rho^i_{t,j} > 0$ for all $i$, $t$, and $j$. Our exogenous state vector for product $i$ at time $t$ is all of this information:
\[
	s^i_t = (D_t^{i}, p_t^{i}, c_t^{i}, U^i_t, \bM^i_t, \boldsymbol\rho^i_t).
\]
To allow for the most general formulation possible, we consider policies that can leverage the history of all products. In our implementation, however, we learn a policy that only uses the history of that product and for our learnability results in \Cref{sec:learn} we will assume independence of the processes between products. Therefore, we will define the history
\[
	H_t :=\{( s_1^i,...,s_{t-1}^i)\}_{i=1}^{|\mathcal{A}|}
\]
as the joint history vector of the external processes for all the products up to time $t$.

\paragraph{Control Processes} Our control process will involve picking actions for each product jointly from a set of all possible actions $\mathbb{A} := \mathbb{R}_{\geq 0}^{|\mathcal{A}|}$. For product $i$, the action taken is denoted by $a_t^{i} \in \mathbb{R}_{\geq 0}$, the order quantity for product $i$. For a class of policies parameterized by $\theta$, we can define the actions as
\[
	a_t^{i} = \pi_{\theta,t}^{i}(H_t).
\]
We characterize the set of these policies as $\Pi = \{\pi_{\theta,t}^{i}| \theta \in \Theta, i \in \mathcal{A}, t \in [0,T]\}.$

\paragraph{Order Quantity Constraints} We allow for the existence of an order quantity post-processer $f_{p} : \RR_{\geq 0} \times \RR^{d_v} \rightarrow \RR_{\geq 0}$ that may modify the order quantity -- for example to satisfy the constraints given by $\bM^{i}_t$. The final order quantity requested from the vendor is denoted as $\tilde{a}^i_t := f_{p}(a_t^i, \bM^{i}_t)$. Note that we do not impose a requirement that any constraints encoded in $\bM_t^i$ be satisfied, merely that we permit the endogenous portion of the state's evolution to depend on such a function as they appear so often in real supply chains.

\paragraph{Inventory Evolution Dynamics}
We assume that the implicit endogenous inventory state follows standard inventory dynamics and conventions. Inventory arrives at the beginning of the time period, so the inventory state transition function is equal to the order arrivals at the beginning of the week minus the demand fulfilled over the course of the week. Both demand and arrivals may be censored due to having lower inventory on-hand or vendor having low supply, respectively. The amount arriving, according to our model of arrivals is:
\begin{equation}
	\label{eq:arrivals}
	I^i_{t_-} = I^i_{t-1} + \sum_{j=0}^L \min(U^i_{t-j},\tilde{a}^i_{t-j})\rho^i_{t-j,j},
\end{equation}
where $I_t^i$ is the inventory at the end of time $t$, and $I_{t_-}^i$ is the inventory at the beginning of time $t$, after arrivals but before demand is fulfilled. Then, at the end of time $t$, the inventory position is:
\[
	I^i_{t} = \min(I^i_{t_-}- D^i_t, 0).
\]

\paragraph{Reward Function} The reward at time $t$ for product $i$ is defined as the selling price times the total fulfilled demand, less the total cost associated with any newly ordered inventory (that  will be charged by the vendor upon delivery):
\begin{equation}
	\label{eqn:rewardfunc}
	R_t^{i} = p_t^i \min(D^i_t, I^i_{t_-}) -  c_t^i \min(U^i_t,\tilde{a}_t^i).
\end{equation}
Note that the cost charged is the {\it realized} order quantity, which is the standard practice in the literature.

We will write $R_t(H_t, \theta)$ to emphasize that the reward is a function only of the exogenous $H_t$ and the policy parameters $\theta$. Recall that selling price and buying cost are determined exogenously. We assume all rewards $R_t^{i} \in [R^{min},R^{max}]$, and assume a multiplicative discount factor of $\gamma \in [0,1]$ representing the opportunity cost of reward to the business. Again, we make the dependence on the policy explicit by writing $R_t^{i}(\theta)$. The objective is to select the best policy (i.e., best $\theta \in \Theta$) to maximize the total discounted reward across all products, expressed as the following optimization problem:
\begin{align*}
	\max_{\theta}       & ~\mathbb{E}_{\mathbb{P}}\Biggl[\sum_{i\in \mathcal{A}} \sum_{t\in [0,T]} \gamma^t R_t^{i}(\theta)\Biggr] \numberthis \label{eq:obj} \\
	\text{subject to: } &                                                                                                                         \\
	I^i_0               & = k^i                                                                                                                   \\
	a^i_t               & = \pi_{\theta,t}^i(H_t)                                                                                                 \\
	\tilde{a}^i_t       & = f_p^i(a^i_t, H_t)                                                                                                     \\
	I^i_{t_-}           & = I^i_{t-1} + \sum_{j=0}^L \min(U^i_{t-j},\tilde{a}^i_{t-j})\rho^i_{t-j,j}, \numberthis \label{eqn:invupdate}           \\
	I^i_{t}             & = \min(I^i_{t_-}- D^i_t, 0).
\end{align*}
Here, $\mathbb{P}$ denotes the joint distribution over the exogenous processes. The inventory $I_0^i$ is initialized at $k_i$, a known quantity \emph{a priori}.

\subsection{Learnability}
\label{sec:learn}

\subsubsection*{Learning Objective}
For the policy to be efficiently learnable, we need to restrict the policy for product $i$ at time $t$ to be a function only of the history of item $i$, $H^i_t :=\{( s_1^i,...,s_{t-1}^i)\}$, and the learnable parameter $\theta$ is shared by all item's policies. The reward is therefore now a function $R_t(H^i_t, \theta)$ of only the history of item $i$ and the parameter $\theta$. The learning objective then becomes
\[
	J_T(\theta) := \EE\Biggl[\sum_{i\in \mathcal{A}} \sum_{t\in [0,T]} \gamma^t R_t^{i}(\theta)\Biggr],
\]
which we estimate via simulation with the objective
\[
	\hat{J}_T(\theta) := \sum_{i\in \mathcal{A}} \sum_{t\in [0,T]} \gamma^t R_t^{i}(\theta).
\]
This is clearly an unbiased estimate of $J_T$ as the historical data $H_T$ is exogenous to the choice of policy.

\subsubsection*{Learnability}
Our problem formulation fits under the framework described in \cite{madeka2022deep}, with additional exogenous variables $U_t$, $\boldsymbol\rho_t$, and $\bM_t$ as part of the external state process. Hence, assuming full observability of these processes, we can accurately simulate the value of any policy.  This follows immediately from Theorem 2\footnote{The addition of the post-processor $f_p$ does not impact the result.} of \citet{madeka2022deep} as we assume that the supply, arrival shares, and vendor constraint processes are exogenous.

In reality, one does not fully observe the supply of the vendors ($U^i_t$) or the arrival shares ($\boldsymbol\rho^i_t$). The supply of the vendor $U_t^i$ is only observed historically at times when the vendor did not fully fill the order -- in which case, $U_t^i = \sum_{j=0}^L o_t^j$. The arrival shares $\boldsymbol\rho^i_t$ are fully observed historically whenever an order is placed and the vendor sends at least one unit.

To proceed we require a way to obtain these missing values. First, for the supply $U_t^i$, the retailer could collect this data by asking vendors to share how many units they are able to supply\footnote{This problem is little studied in the literature, but in some scenarios it is known that the supplier is always at least as well off if they share capacity information \citep{bao2006supply}}.  Another approach is to treat this as a missing data problem, and then use additional exogenous observed context $x_t^i \in \RR^D$ that is available at time $t$ for product $i$, to forecast these unobserved components. The observed context history is denoted as  $X_T^i := (x_1^i,\dots,x_T^i)$. The latter is similar to the uncensoring of demand in \citet{madeka2022deep}.

\begin{assumption}[Accurate Forecast of Supply and Arrival Shares]
	\label{assm:fill-supply}
	Let $H^i_{T,F} := (U^i_1,\boldsymbol\rho^i_1,\dots,U^i_T,\boldsymbol\rho^i_T)$ denote the history of the unobserved exogenous supply and arrival shares processes through time $T$. Likewise, denote the observed components of the exogenous history $H^i_T$ as $H^i_{T,O}$. Now, we can consider the distributions $\PP_F^i := \PP(H^i_{T,F} | H^i_{T,O}, X^i_T)$ and $\hat{\PP}_F^i := \hat{\PP}(H^i_{T,F} | H^i_{T,O}, X^i_T)$. If
	\[
		\frac{1}{|\cA|}\sum_{i\in \cA} ||\hat{\PP}_F^i, \PP_F^i||_{TV} \leq \epsilon_F,
	\]
	we call $\hat{\PP}_F^i$ an accurate forecast of $ \PP_F^i$.
\end{assumption}

Under \Cref{assm:fill-supply}, it follows from Theorem 3\footnotemark[2] of \citet{madeka2022deep} that the inventory control problem with general arrivals is efficiently learnable in the case where we do not observe the supply and arrival shares processes. In practice, we may choose to forecast {\it arrivals} instead of the supply and arrival shares processes -- see \Cref{rem:fcst-arrivals} below.

\begin{remark}[Forecasting Arrivals]
	\label{rem:fcst-arrivals}
	Note that the dynamics \eqref{eqn:invupdate} and reward function \eqref{eqn:rewardfunc} depend only on the arrivals $o_{t,j}^i := \min(U^i_t,f_p(a_t^i,\bM^i_t))\rho^i_{t,j}$, so we forecast arrivals conditional on the action $a_t^i$ rather than the supply and arrival share processes for the purposes of constructing our simulator from historic data.
\end{remark}

\subsection{Modeling arrivals with Gen-QOT}
\label{sec:gen-qot}
Having established that our problem of interest is efficiently learnable, we proceed with describing the QOT model and then evaluating the model. Per \cref{rem:fcst-arrivals}, we forecast the arrival sequence directly rather than the supply and arrival processes. Formally, we forecast the distribution
\[
	p(o^i_{t,0},\dots, o^i_{t,L} | H_{t,O}^i, X_t^i).
\]
The model is trained to minimize log-likelihood of the forecasted distribution. See \Cref{sec:genqot} for a complete description of the model and training objective. It is worth emphasizing that modeling arrivals directly allows us to treat $f_p$ as part of the forecast of the arrivals process, which will be advantageous in \Cref{sec:method-sim} when we describe how to make the simulator differentiable.

\begin{figure}[!htb]
	\centering
	\includegraphics[width=0.49\textwidth]{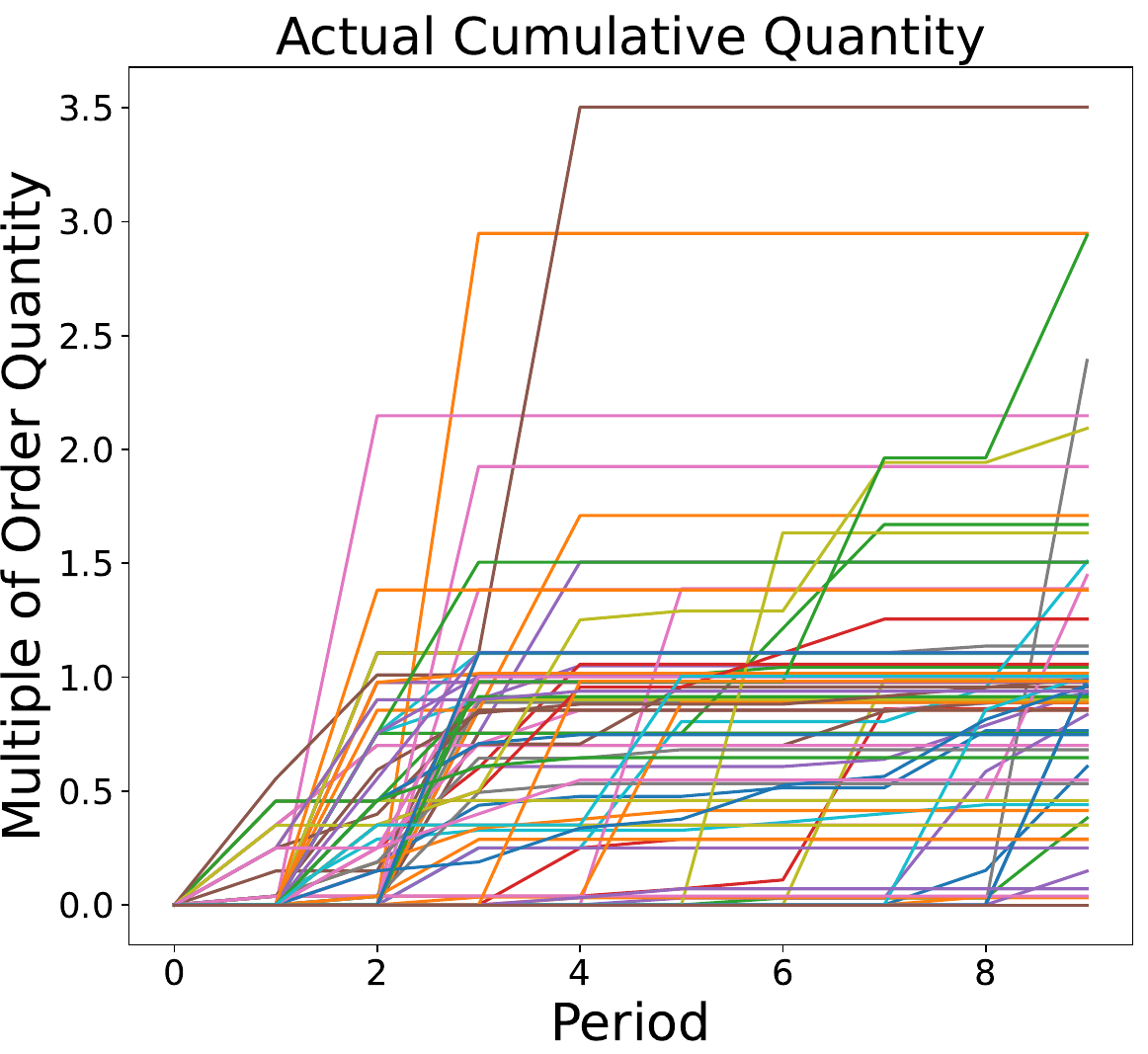}
	\includegraphics[width=0.49\textwidth]{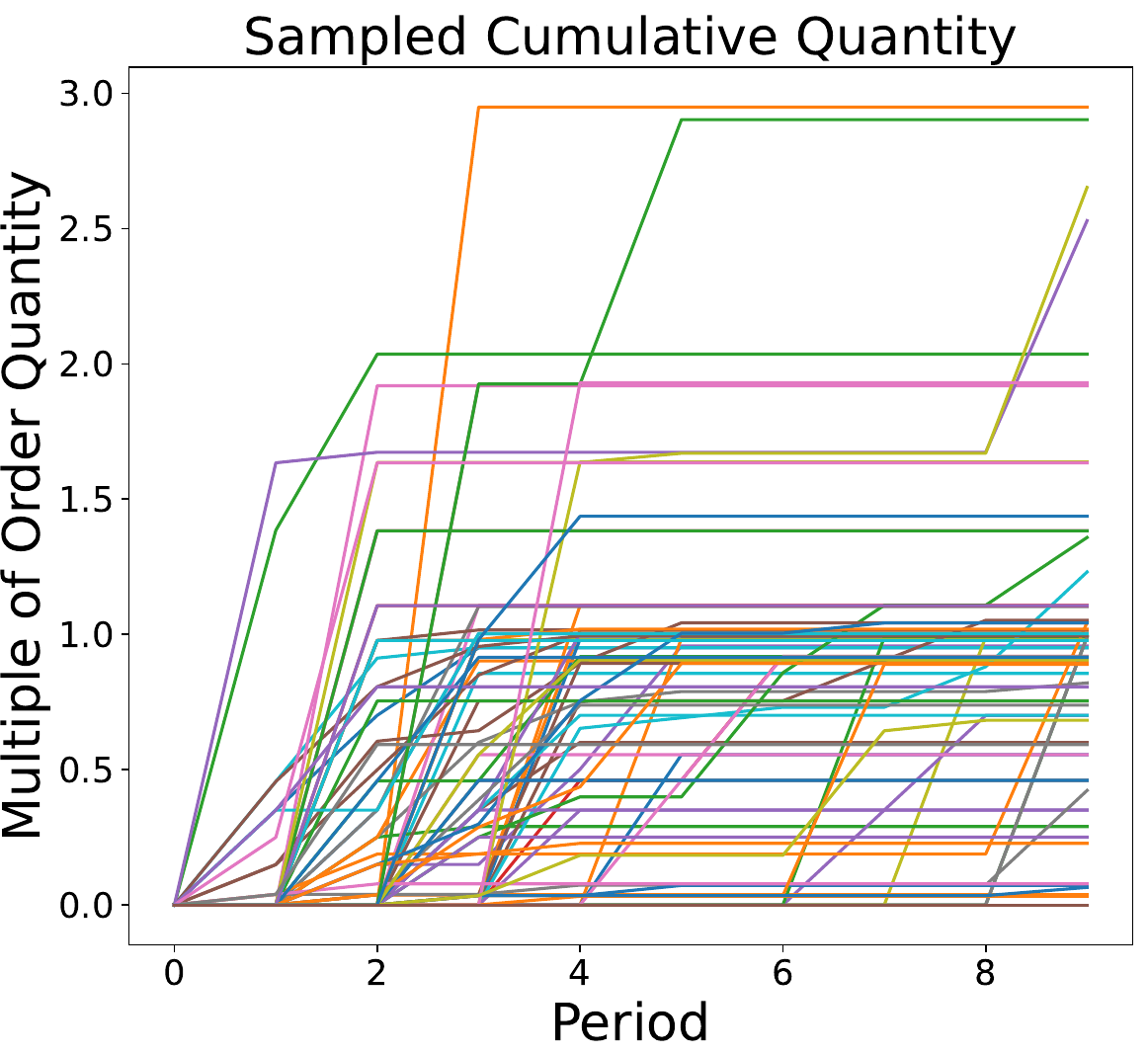}
	\caption{A set of 256 real and simulated sample paths of order-quantity normalized inventory arrivals.}
	\label{fig:rl_qot_sample_paths}
\end{figure}

\Cref{fig:rl_qot_sample_paths} shows a set of sample paths generated from our Gen-QOT model alongside a set of real order-quantity normalized inventory arrivals.
A natural question then arises: how should one measure the quality of the generated sample paths? The simplest thing would be to compare against other methods for forecasting vendor lead times (this will be referred to as {\bf Criterion 1}). In addition, we might want to check that the generated paths satisfy several desirable properties that we anticipate may be relevant to the inventory control problem:
\begin{enumerate}
	\item {\bf Criterion 2} -- Does Gen-QOT predict the right amount of cumulative inventory $l$ weeks after an order is placed?
	\item {\bf Criterion 3} -- Does Gen-QOT predict receiving zero inventory for the correct orders?
	\item {\bf Criterion 4} -- Does Gen-QOT predict correctly whether there is an arrival in the first week after an order is placed?
	\item {\bf Criterion 5} -- Does Gen-QOT predict correctly the time by which the order fully arrives?
\end{enumerate}
Accordingly, we propose five methods for evaluating the quality of the dynamics model:
\begin{enumerate}
	\item {\bf Criterion 1} -- Obtain the empirical distribution predicted by Gen-QOT and evaluate against with standard accuracy metrics such as CRPS and quantile loss.
	\item {\bf Criterion 2} -- Regress the actual cumulative inventory received on the mean predicted cumulative inventory received for each week.
	\item {\bf Criterion 3 and 4} -- Generate sequences and use as a classifier, then construct a classifier calibration plots for generated sequences.
	\item {\bf Criterion 5} -- Generate sequences and use them to generate probabilistic forecasts of the arrival time of the full order, then compute the calibration.
\end{enumerate}
In \Cref{sec:eval-dynamics} we fit the Gen-QOT model to historic data and perform evaluations against the five criterion listed above.

\subsection{Simulator Construction}
\label{sec:method-sim}
Our simulator is constructed similarly to \citet{madeka2022deep} aside from the change to the inventory transition dynamics. To handle the general arrival dynamics, the first approach is to estimate the partially observed $H^{i}_{t,F}$ and directly implement \eqref{eqn:invupdate} since $f_p$ is known. Alternatively, following \Cref{rem:fcst-arrivals}, we could forecast arrivals and simply sample that distribution at each step. The issue with both these approaches are that the simulator is no longer path-wise differentiable and if possible, we would prefer to leverage the fact that most of our dynamics {\it are differentiable} and thus the exact gradient can be computed analytically.

The approach we take in our empirical work is the following: first, given the action $a_t$ and the exogenous $H_{t,O}^{i}$, $X_t^i$, sample the estimated forward model
\begin{equation}
	\label{eqn:arrival-sampling}
	\hat{o}^i_{t,0},\dots, \hat{o}^i_{t,L} \sim \hat{p}(\cdot | H_{t,O}^{i}, X_t^i, a_t).
\end{equation}
The sampled arrivals can then be converted into a sampled sequence of partial fills $\bar{\alpha}^i_{t,l} = \frac{\hat{o}^i_{t,l}}{a^i_{t}}$ by rescaling by the action $a_t^i$. The inventory update in \eqref{eqn:invupdate} becomes
\begin{equation}
	\label{eqn:invupdate-sampled}
	I^i_{t_-} = I^i_{t-1} + \sum_{j=0}^L \bar{\alpha}^i_{t-j,j} a^i_{t-j},
\end{equation}
This is similar in spirit to the approach in \citet{clavera2020model}, except they use the re-parametrization trick and then differentiate through the forward model, passing a noise process as input.

\section{Empirical Results}
\label{sec:eval}
In this section we present some empirical results. First, in \Cref{sec:eval-dynamics} we evaluate the performance of the fitted Gen-QOT under the four criterion discussed in \Cref{sec:gen-qot}. Then in \Cref{sec:eval-sim} we demonstrate through backtests against historical data in a simulator based on the Gen-QOT model the difference in performance between baseline policies, an RL trained in a simulator with Gen-QOT dynamics and an RL under dynamics dictated by a classical vendor lead-time model. Finally, in \Cref{sec:eval-real} we show recent results from a real-world A/B test of the RL policy in the US store of a large e-retailer.

\subsection{Evaluating the Gen-QOT Model}
\label{sec:eval-dynamics}
First, we evaluate our proposed Gen-QOT model as we require an accurate forecast in order for the policy backtest to be valid. The Gen-QOT model architecture and training objective can be found in \Cref{sec:genqot}.

\subsubsection*{Training and Evaluation Data}
We train Gen-QOT on purchase orders from 250K products from the US store of a large e-retailer from 2017-05-13 to 2019-02-01 and holdout 100K actions from 2019-02-01 to 2020-02-01 to evaluate model performance. This time period allows us to judge both in-time and out-of-time time generalization. The features used in our model can be found in \Cref{sec:featurization}.

\subsubsection*{Results}
\paragraph{VLT Forecasting} Because many inventory control systems rely on simplified optimization models that assume only random lead-time, we evaluate the Gen-QOT model against one that directly predicts quantiles of the vendor lead-time distribution. The architecture used is similar to Gen-QOT, but we replace the recurrent neural-net decoder with a simple multi-layer perceptron.\footnote{More precisely, the architecture is MQCNN which achieve SOTA performance on probabilistic forecasting tasks that use very similar data -- see ``MQ\_CNN\_wave'' in Figure 3 of \citet{wen2017mqcnn}. }

To produce a forecast of the vendor lead time from Gen-QOT, samples are generated from Gen-QOT to obtain an empirical distribution from which quantiles can be determined. \Cref{table:genq-qot-vs-direct} shows the backtest results on data from 2019-02-01 to 2020-02-01, showing that Gen-QOT is competitive with traditional vendor lead time forecasting approaches. See \Cref{sec:metrics} for definitions of the CRPS and Quantile Loss metrics used.

\begin{table}[H]
	\caption{Backtest of generative model versus direct quantile forecast (lower values are better); 95\% confidence intervals are on the performance gap between the two models.}
	\begin{center}
		\begin{tabular}{ c c c c  }
			\toprule
			       & \multicolumn{2}{c}{Model} &                         \\
			\cmidrule{2-3}
			Metric & Direct Prediction         & Gen-QOT & 95\% CI       \\
			\midrule
			CRPS   & 100.00                    & 101.61  & [-0.23, 3.46] \\
			P10 QL & 100.00                    & 99.92   & [-1.91, 1.75] \\
			P30 QL & 100.00                    & 101.24  & [-0.57, 3.04] \\
			P50 QL & 100.00                    & 102.41  & [0.38, 4.44]  \\
			P70 QL & 100.00                    & 103.21  & [0.89, 5.54]  \\
			P90 QL & 100.00                    & 102.22  & [0.50, 4.96]  \\
			\bottomrule
		\end{tabular}
	\end{center}
	\label{table:genq-qot-vs-direct}
\end{table}

See \Cref{sec:neural-ablation} for an ablation study across different candidate architectures for Gen-QOT comparing their performance under the CRPS and Quantile Loss metrics.

\paragraph{Calibration and Classifier Metrics} Next we check the calibration of the cumulative receives forecast and arrival time forecasts that can be inferred from Gen-QOT -- see \Cref{sec:metrics-cal} for how we define calibration in general, and \Cref{sec:arrival-time-cal} for arrival time calibration. \Cref{table:rl-qot-dbp-calibration} shows the calibration of Gen-QOT's forecasted distributions of cumulative arrivals. Most coefficients are close to one, showing that Gen-QOT predicts the cumulative quantity of inventory received for a specific order over time reasonably well, although it is not perfectly calibrated (coefficient of 1).

\begin{figure}[h]
	\centering
	\begin{subfigure}{0.4\textwidth}
		\includegraphics[width=\textwidth]{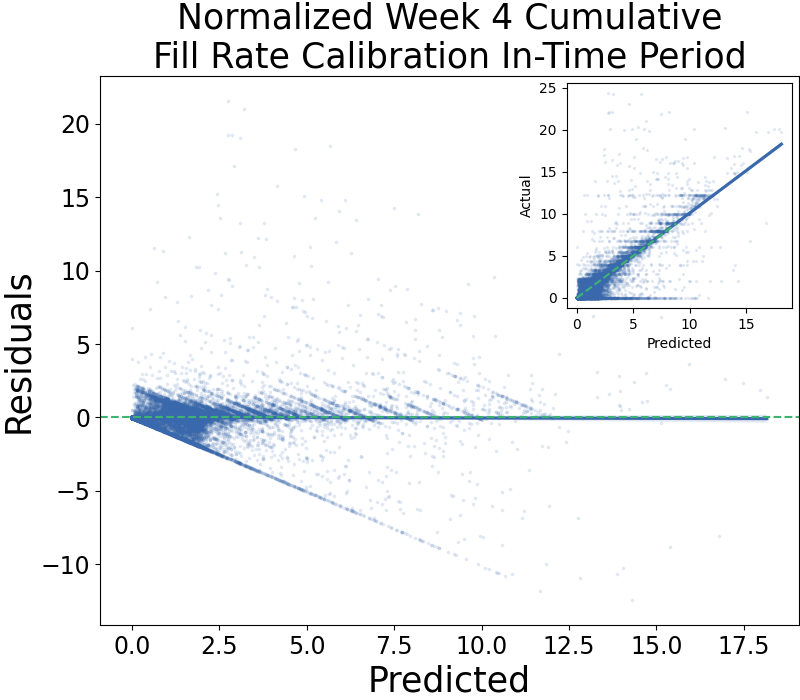}
		\caption{}
		\label{fig:cumulative_recieve_week4_calibration_intime}
	\end{subfigure}
	\begin{subfigure}{0.4\textwidth}
		\includegraphics[width=\textwidth]{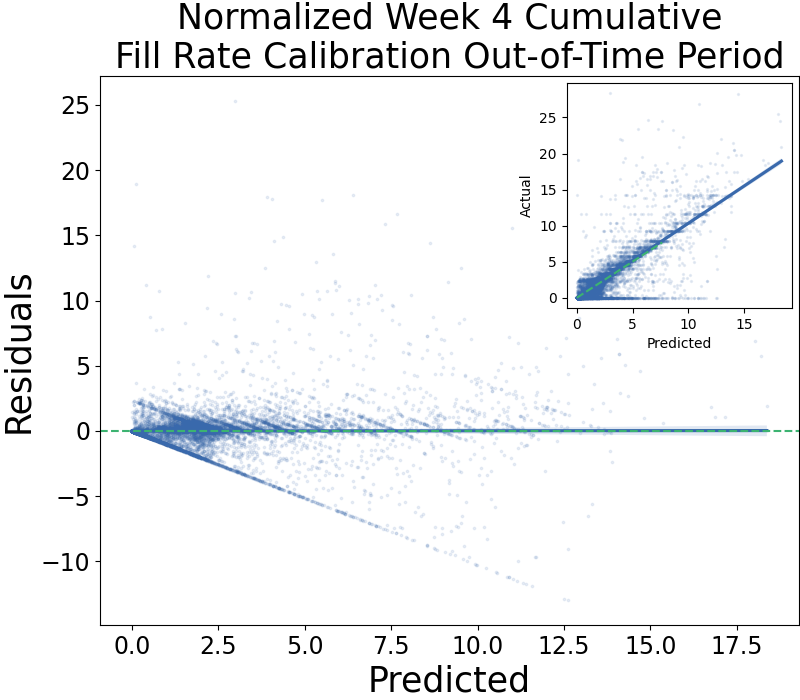}
		\caption{}
		\label{fig:cumulative_recieve_week4_calibration_outoftime}
	\end{subfigure}
	\caption{Residual calibration plot for cumulative inventory received -- residuals are plotted against predicted values in the main figure, while the original plot is shown in an inner figure.}
	\label{fig:cumulative_recieve_week4_calibration}
\end{figure}

\begin{table}[h]
	\caption{Calibration of cumulative inventory received predicted $k$ weeks after submitting an order and actuals.}
	\label{table:rl-qot-dbp-calibration}
	\begin{center}
		\begin{tabular}{ccccc}
			\toprule
			\multicolumn{1}{c}{} & \multicolumn{2}{c}{In Time Holdout} & \multicolumn{2}{c}{Out of Time Holdout}                               \\
			\cmidrule{2-5}
			Weeks After Order    & Estimate                            & 95\% CI                                 & Estimate & 95\% CI          \\
			\midrule
			1                    & 1.0577                              & [1.056,  1.060]                         & 1.0559   & [1.055,  1.057]  \\
			2                    & 1.1393                              & [1.138,   1.141]                        & 1.1529   & [1.151,  1.154]  \\
			3                    & 1.108                               & [1.107,     1.109]                      & 1.1311   & [1.130,   1.132] \\
			4                    & 1.0866                              & [1.086,    1.087]                       & 1.1147   & [1.114,  1.115]  \\
			5                    & 1.0789                              & [1.078,      1.079]                     & 1.1094   & [1.109,   1.110] \\
			6                    & 1.0745                              & [1.074,    1.075]                       & 1.1021   & [1.102,   1.103] \\
			7                    & 1.0694                              & [1.069,    1.070]                       & 1.0995   & [1.099,   1.100] \\
			8                    & 1.063                               & [1.063,    1.063]                       & 1.0953   & [1.095,   1.096] \\
			9                    & 1.0538                              & [1.053,    1.054]                       & 1.0895   & [1.089,   1.090] \\
			\bottomrule
		\end{tabular}
	\end{center}
\end{table}
\Cref{fig:cumulative_recieve_week4_calibration} shows this on samples of actual vs. predicted mean for normalized cumulative inventory received at the end of the fourth week on the held out orders from the training period and the test period. The results on the out-of-time hold out set are very similar to those in the training period.

For arrival time calibrations, intuitively, we are measuring: if the forecaster predicts with 25\% chance by a specific date, does it arrive by that date 25\% of the time? \Cref{tab:arrival-time-cal-backtest} shows that on both the in-time and out-of-time holdouts, the arrival time forecasts generated by Gen-QOT are well calibrated.  See \Cref{sec:arrival-time-cal-results} for full arrival time calibration metrics, further broken out by lead time.

\begin{table}[H]
	\caption{Calibration of arrival time for samples from two holdout sets, by forecasted probability}
	\begin{center}
		\begin{tabular}{cccccccc}
			\toprule
			\multicolumn{1}{c}{} & \multicolumn{3}{c}{In Time Holdout} & \multicolumn{3}{c}{Out of Time Holdout}                               \\
			\cmidrule{2-7}
			Probability & Avg. Pred. & Average & 95\% CI & Avg. Pred. & Average & 95\% CI \\
			\midrule
			0.0-0.1         &           0.04 &        0.05 &   [0.05, 0.05]   	&	  0.04 &        0.05 &   [0.05, 0.05]   \\
			0.1-0.2         &           0.14 &        0.15 &   [0.15, 0.15]   	&	  0.14 &        0.14 &   [0.14, 0.14]   \\
			0.2-0.3         &           0.25 &        0.24 &   [0.24, 0.24]   	&	  0.25 &        0.24 &   [0.24, 0.24]   \\
			0.3-0.4         &           0.35 &        0.34 &   [0.34, 0.34]   	&	  0.35 &        0.33 &   [0.33, 0.33]   \\
			0.4-0.5         &           0.45 &        0.43 &   [0.43, 0.43]   	&	  0.45 &        0.43 &   [0.43, 0.43]   \\
			0.5-0.6         &           0.55 &        0.54 &   [0.53, 0.54]   	&	  0.55 &        0.53 &   [0.53, 0.53]   \\
			0.6-0.7         &           0.65 &        0.64 &   [0.64, 0.64]   	&	  0.65 &        0.63 &   [0.63, 0.63]   \\
			0.7-0.8         &           0.75 &        0.74 &   [0.74, 0.74]   	&	  0.75 &        0.74 &   [0.74, 0.74]   \\
			0.8-0.9         &           0.85 &        0.84 &   [0.84, 0.84]   	&	  0.85 &        0.84 &   [0.84, 0.85]   \\
			0.9-1.0         &           0.99 &        0.99 &   [0.99, 0.99]   	&	  0.99 &        0.99 &   [0.99, 0.99]   \\
			\bottomrule
		\end{tabular}
		\label{tab:arrival-time-cal-backtest}
	\end{center}
\end{table}

The classifier calibration plots are produced by discretizing and binning the predicted probabilities of the event we are interested in, and then estimating the mean of the actual classifications for each bin.  Ideally, the average actual will be equal to the mean of the predicted probabilities in each bin, and this point will fall along the 45$^{\circ}$ line.
\begin{figure}[h]
	\centering
	\begin{subfigure}{0.45\textwidth}
		\includegraphics[width=\textwidth]{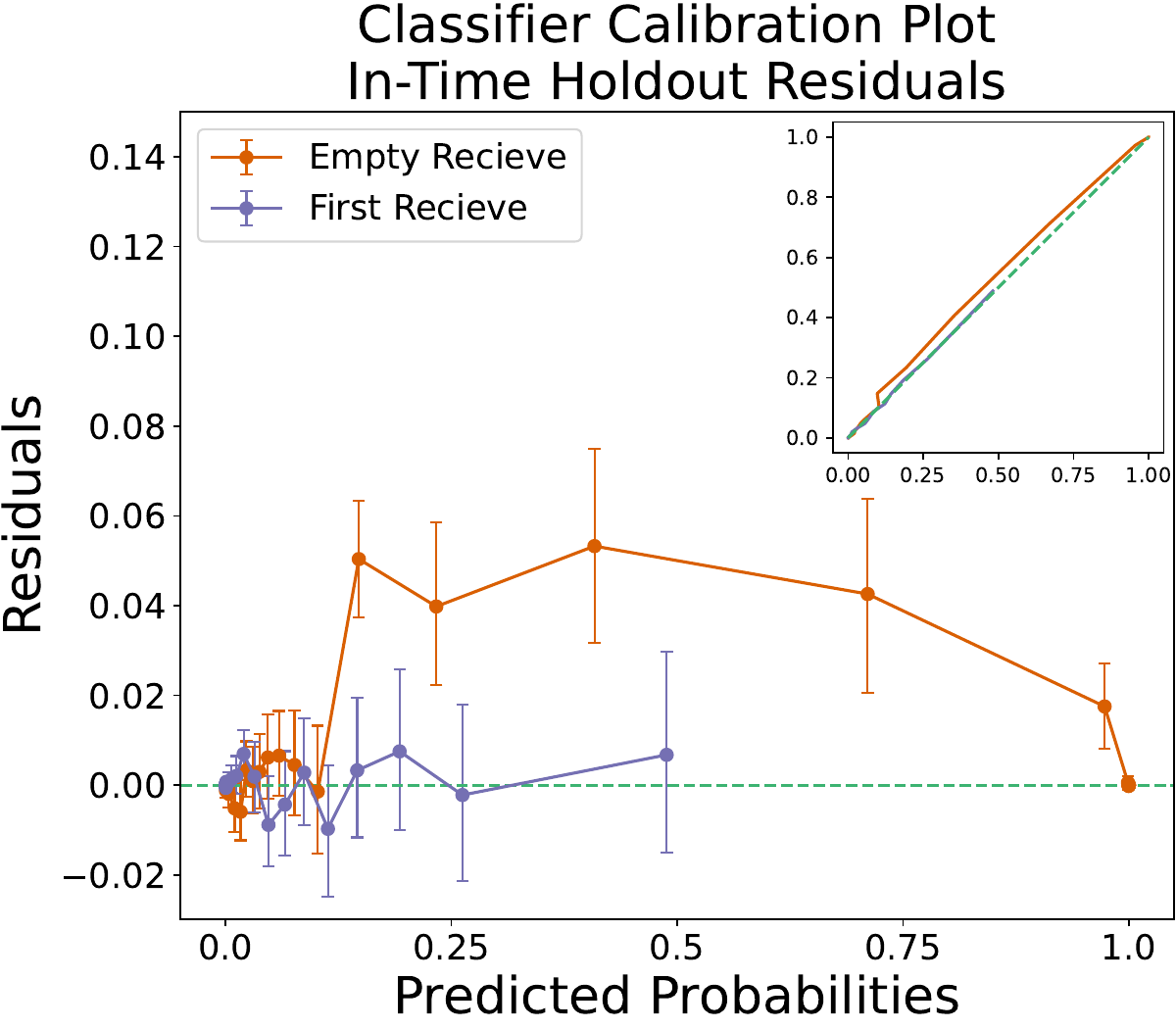}
		\caption{}
		\label{fig:empty_calibration_intime}
	\end{subfigure}
	\begin{subfigure}{0.45\textwidth}
		\includegraphics[width=\textwidth]{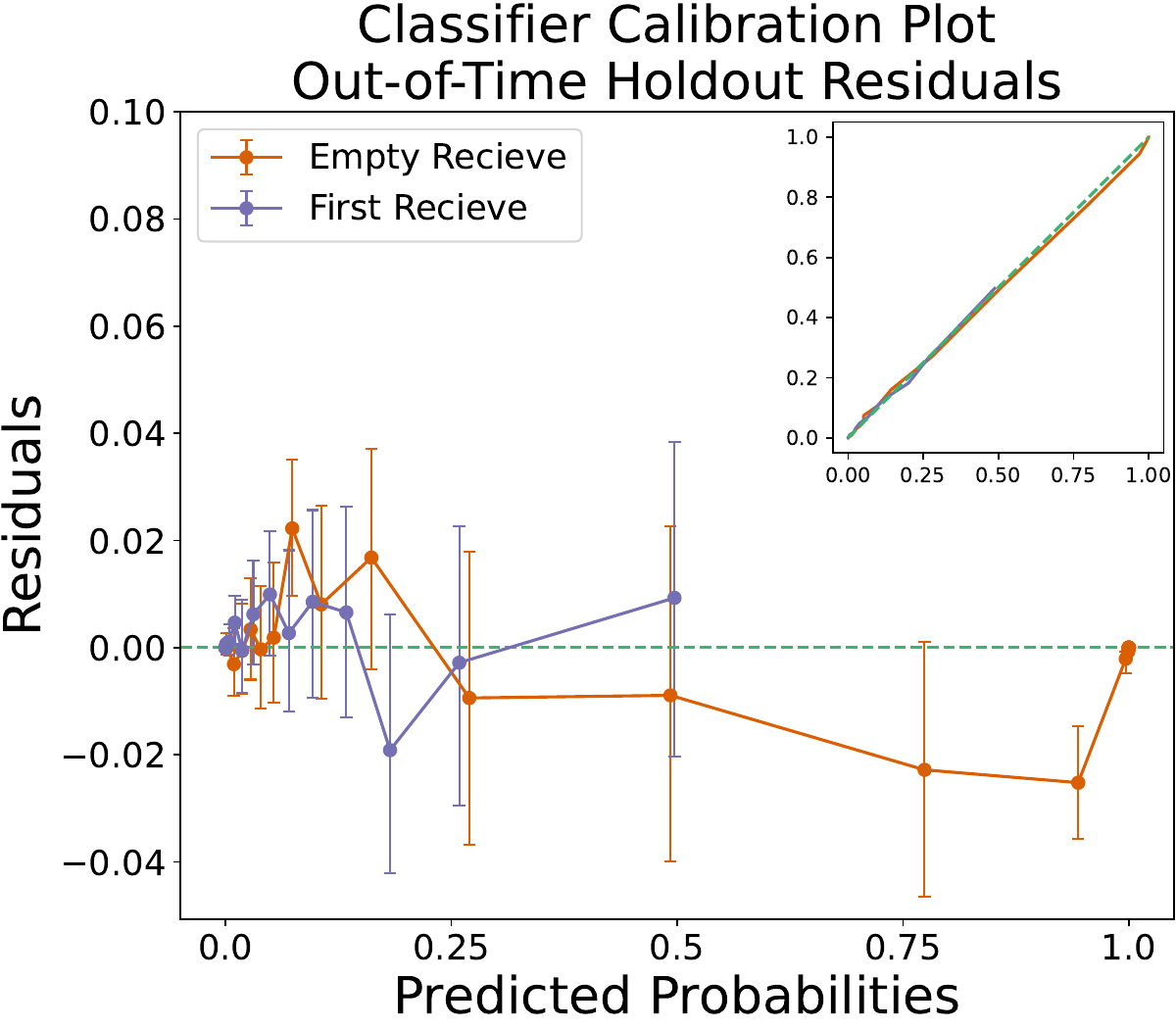}
		\caption{} 
		\label{fig:empty_calibration_outoftime}
	\end{subfigure}
	\caption{Residual calibration plots for Gen-QOT -- residuals are plotted against predicted values in the main figure, while the original plot is shown in an inner figure.}
	\label{fig:calibration-metrics}
\end{figure}
\Cref{fig:calibration-metrics} shows the calibration of Gen-QOT at predicting whether a purchasing action will yield zero inventory for both the in-time holdout (\Cref{fig:empty_calibration_intime}) and out-of-time holdout (\Cref{fig:empty_calibration_outoftime}). In both cases, points generally follow the ideal calibration line, with some slight and expected degradation in the out-of-time holdout.

\subsubsection*{Model Behavior}
\begin{figure}[h]
	\centering
		\includegraphics[width=0.49\textwidth]{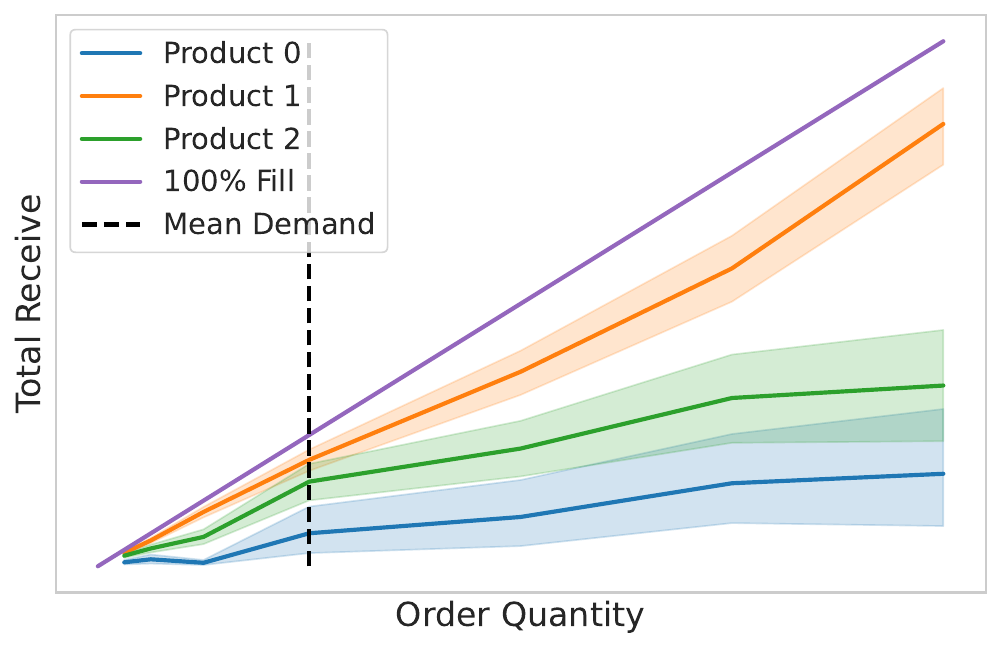}
	\caption{QOT predicted receive versus orders}
	\label{fig:qot-counterfactuals}
\end{figure}

Figure \Cref{fig:qot-counterfactuals} shows the response of the mean end-to-end yield predicted by the QOT model to varying order quantities for several randomly selected products. As we can see, for some products the model predicts nearly a full yield regardless of order quantity, while for others, the yield tapers off as the order quantity increases.

\subsection{Backtest of Inventory Control Policies}
\label{sec:eval-sim}
Having demonstrated in \Cref{sec:eval-dynamics} that we have a dynamics model that achieves good accuracy on our dataset, the next thing we want to do is use our generative model to backtest various policies and measure their performance.

\subsubsection*{Data}
We use a similar dataset to that used to fit the Gen-QOT model. The training period is data from 2017-05-13 to 2019-02-01 and the backtest period is  2019-02-01 to 2020-02-01. The features used for the policy and the simulator can be found in \Cref{sec:featurization}.

\subsubsection*{Policies  and Training}
Because \citet{madeka2022deep} performed an exhaustive evaluation of various policies, we consider only a base stock baseline and two RL policies -- one trained on a gym that uses the inventory arrival dynamics from \citet{madeka2022deep} and one trained on gym with Gen-QOT arrival dynamics. These will be referred to as {\bf Newsvendor}, {\bf VLT-DirectBP}, and {\bf QOT-DirectBP}, respectively. The policy networks consist of a Wavenet encoder \citep{oord2016wavenet} and MLP decoder.

Following \citet{madeka2022deep} we implement a differentiable simulator in PyTorch. All algorithms are trained using a single p3dn.24xlarge EC2 instance. We use the DirectBackprop algorithm to train the {\bf VLT-DirectBP} and {\bf QOT-DirectBP} agents.

\subsubsection*{Results}
\Cref{table:rl-qot-dbp-results} summarizes the changes in the sum of discounted reward of both policies relative to a baseline policy. QOT-DirectBP outperforms VLT-DirectBP when evaluated under the QOT transition dynamics. While it is unsurprising that the policy trained on the simulator (with Gen-QOT) used in evaluation performs best, this does underscore the impact of a ``Sim2Real'' gap -- in this case ``reality'' is the QOT simulator -- as the overall performance gain is ~8\%. Both policies still outperform the Newsvendor baseline, which is unsurprising given the results in \citet{madeka2022deep}.

\begin{table}[h]
	\caption{Comparison of RL policies in a backtest using Gen-QOT. 95\% confidence intervals are on the difference from baseline. }
	\begin{center}
		\begin{tabular}{ c c c}
			\toprule
			Policy              & Discounted Reward & 95\% CI             \\
			\midrule

			Newsvendor Baseline & 100.00\%          & --                  \\
			VLT-DirectBP        & 109.64\%          & [8.83\% , 10.45\%]  \\
			QOT-DirectBP        & \textbf{117.81}\% & [16.92\% , 18.69\%] \\

			\bottomrule
		\end{tabular}
	\end{center}
	\label{table:rl-qot-dbp-results}
\end{table}

\begin{figure}[h]
	\centering
	\begin{subfigure}{0.4\textwidth}
		\includegraphics[width=\textwidth]{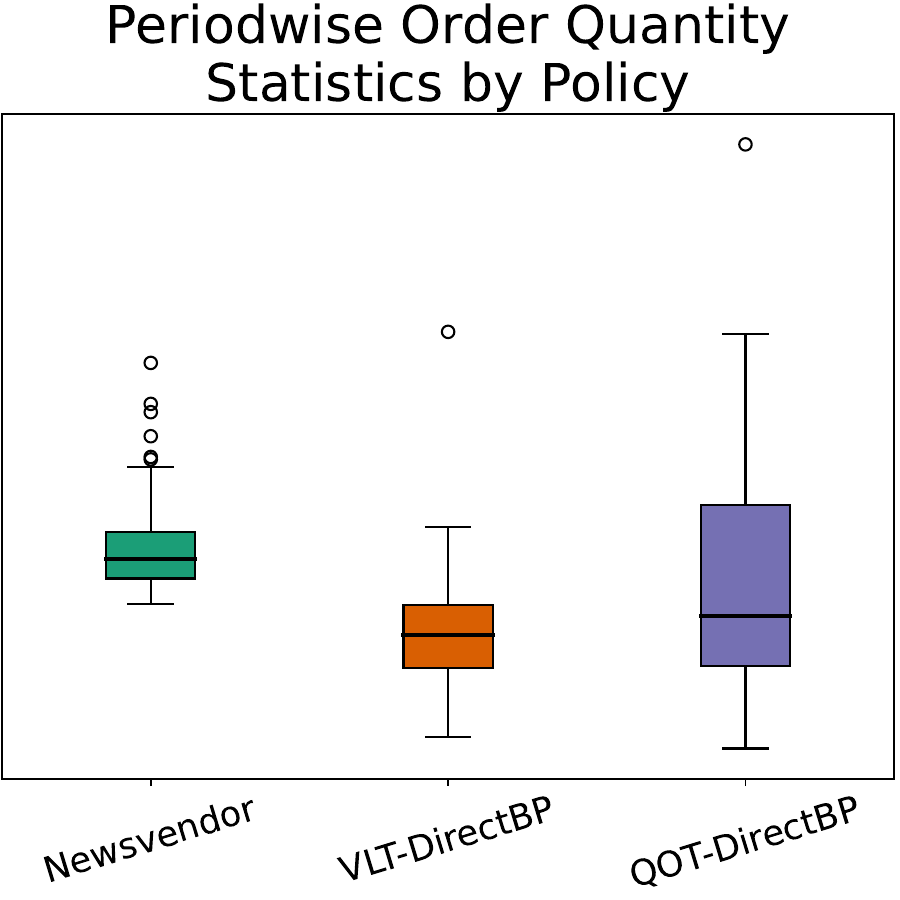}
		\caption{}
	\end{subfigure}
	\begin{subfigure}{0.4\textwidth}
		\includegraphics[width=\textwidth]{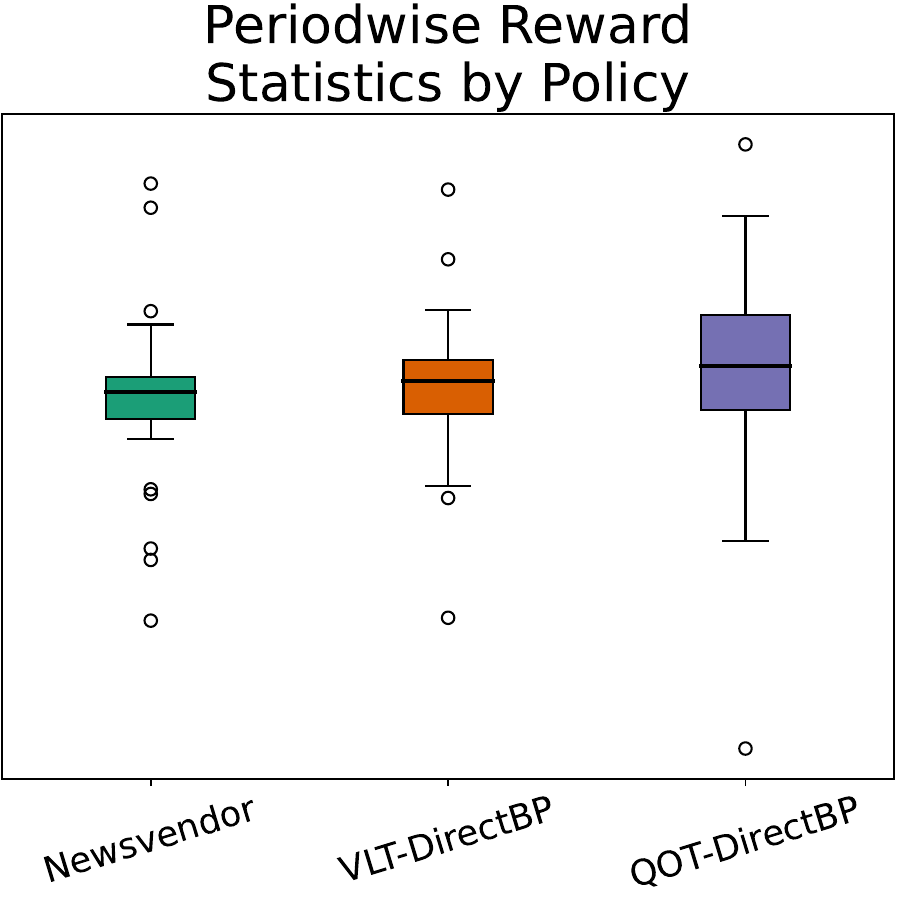}
		\caption{}
	\end{subfigure}
	\caption{Distribution of per-period statistics for order quantity and rewards in the backtest period (2019-02 to 2020-02).}
	\label{fig:per-period-stats}
\end{figure}

\begin{figure}[h]
	\centering
		\includegraphics[width=0.5\textwidth]{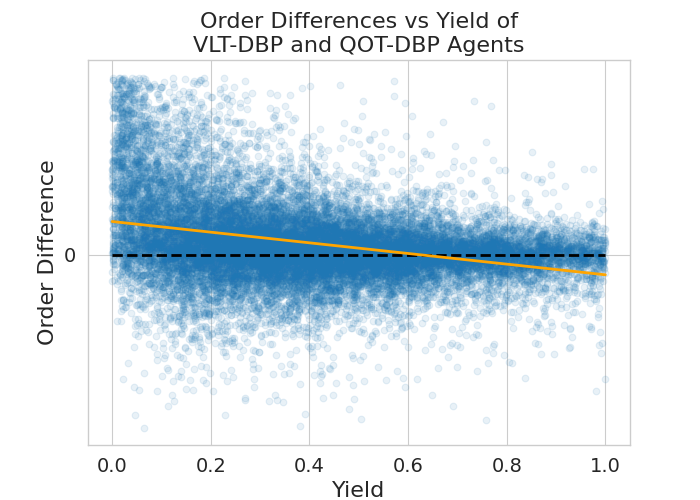}
	\caption{Comparison of difference in mean order quantity between the QOT-DirectBP and VLT-DirectBP agents versus yield; the orange line shows the OLS regression line fit to the data. }
	\label{fig:gym-order-comparison}
\end{figure}

In addition to cumulative discounted reward, we can also consider the distribution of period-wise statistics. \Cref{fig:per-period-stats}. What is interesting is that QOT-DirectBP selects larger order quantities, which likely reflects the fact that Gen-QOT captures stochastic yields. We also note that QOT-DirectBP has higher mean and median reward than other policies. Indeed, \Cref{fig:gym-order-comparison} shows that for lower yield products, the mean order placed by QOT-DirectBP is higher than that of VLT-DirectBP. Put differently -- the QOT-DirectBP policy can adjust to variable yield rates at the product level.


\subsection{Real World A/B Test}
\label{sec:eval-real}

Finally, we ran A/B tests comparing RL inventory control policies to the existing production policy (a base stock policy) at a large e-retailer lasting several months and covering thousands of products. \Cref{table:ab-te} summarizes the treatment effect of all tests on several quantities of interest: reward, inventory level, order quantity and sales. Trial 1 evaluated the performance of the VLT-DirectBP policy while Trials 2 and 3 evaluated the QOT-DirectBP policy.

\begin{table}[H]
	\caption{Treatment effect estimate (percent change) on reward, inventory level, and order quantity in real world A/B tests -- all results shown are significant at the 95\% confidence level.}
	\begin{center}
		\begin{tabular}{c c c c}
			\toprule
			Quantity       & Trial 1 & Trial 2 & Trial 3         \\
			\midrule
			 Reward          & $\sim$ & 3.5\% & 2.7\% \\
			 Inventory Level & -15.2\% &  11.1\% & 3.3\% \\
			 Order Quantity  &  31.1\% &  -7.2\% & $\sim$           \\
			 Sales 			 & $\sim$ &  3.4\% &  1.8\%\\
			\bottomrule
		\end{tabular}
	\end{center}
	\label{table:ab-te}
\end{table}

\subsubsection*{Trial 1: VLT-DirectBP}
\begin{figure}[h]
	\centering
		\includegraphics[width=0.5\textwidth]{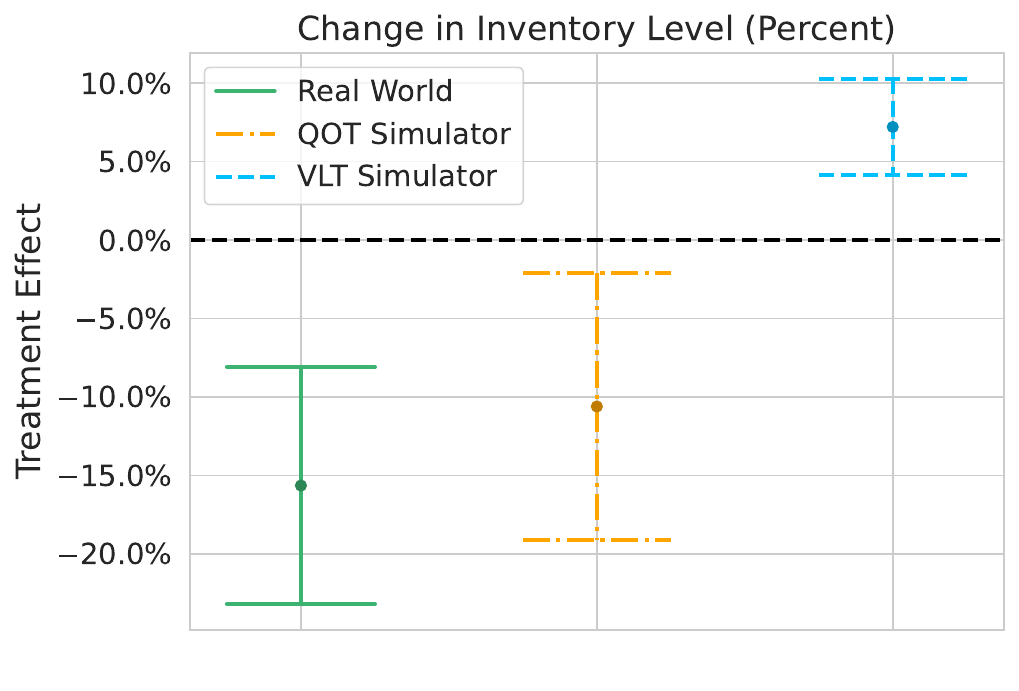}
	\caption{Treatment effect on inventory level of VLT-DirectBP: estimated on rollouts in real world, QOT simulator and VLT simulator.}
	\label{fig:treatment-effect-real-qot-vlt}
\end{figure}

In the first trial, we used the VLT-DirectBP policy described above as the treatment\footnote{This is the same trial described in \citet{madeka2022deep}}. \Cref{fig:treatment-effect-real-qot-vlt} shows the treatment effect estimate on inventory level observed in the actual A/B test alongside the treatment effect estimate from rollouts using the {\it same data} in the QOT and VLT based simulators. We see that the point estimate of the treatment effect in the real supply chain is contained in the confidence interval. This suggests that -- at least as it pertains to inventory level -- the QOT based simulator captures what happens in the actual supply chain.

\subsubsection*{Trial 2 and Trial 3: QOT-DirectBP}
Next, we ran two randomized control trials of the QOT-DirectBP agent in the US store of a large e-retailer lasting several months and covering thousands of products. The control arm used the existing production system (a base stock policy). The results of both these tests can be found in \Cref{table:ab-te}. In Trial 2, the inventory and reward increase by a statistically significant amount -- demonstrating that RL policies {\it can} outperform sophisticated base stock policies in real world settings. In Trial 3, we deployed a QOT-DirectBP policy that we expected (based on backtests) to hold a similar amount of inventory to the existing production system. 

\subsubsection*{Gen-QOT Performance ``off-policy''}

We use data from the Treatment and Control arms of the third A/B test described above to validate our assumption on the forecast accuracy of the QOT model. We already know from \Cref{sec:eval-dynamics} that the QOT model generalizes well out of sample and forward in time. But the question remains: does it generalize well out-of-sample {\it and off-policy}? This is critical because in order for our {\it inventory control backtest} to be accurate, we required \Cref{assm:fill-supply}.

To validate our assumption that QOT does generalize off-policy, we check the forecast errors on the treatment arm versus the control arm.  In \Cref{tab:genqot-offpolicy-cal} and  \Cref{tab:genqot-offpolicy-ql} we see that the difference in forecast performance on-policy versus off-policy in the actual supply chain is not statistically significant.

\begin{table}[H]
	\centering
	\begin{subfigure}[b]{0.58\textwidth}
		\begin{center}
			\begin{tabular}{ccc}
				\toprule
				 &\multicolumn{2}{c}{Change (Control - Treatment)} \\
						\cmidrule{2-3}
				Fcst. Probability & Mean &    95\% CI \\
				\midrule
				0.0-0.1 &   4.61\% & [-0.88\%, 10.11\% ] \\
				0.1-0.2 &   3.17\% & [-0.98\%, 7.34\% ]\\
				0.2-0.3 &   2.62\% & [-1.86\%,  7.12\%] \\
				0.3-0.4 &   2.43\% & [-1.18\%,  6.05\%] \\
				0.4-0.5 &   3.25\% & [0.89\%,  5.61\%] \\
				0.5-0.6 &   1.43\% & [-0.25\%,  3.13\%] \\
				0.6-0.7 &   0.17\% & [-1.57\%,  1.92\%] \\
				0.7-0.8 &  -0.10\% & [-1.59\%,  1.37\%] \\
				0.8-0.9 &   0.01\% & [-0.99\%,  1.02\%] \\
				0.9-1.0 &  -0.01\% & [-0.21\%,  0.18\%] \\
				\bottomrule
				\end{tabular}
			\label{tab:arrival-time-cal-labs}
		\end{center}
		\caption{Calibration of arrival times, by forecast probability}
	\end{subfigure}
	\begin{subfigure}[b]{0.41\textwidth}
		\begin{center}
			\begin{tabular}{ccc}
				\toprule
				    & \multicolumn{2}{c}{Change (Control - Treatment)}                       \\
				\cmidrule{2-3}
				$k$ & Mean                                             & 95\% CI             \\
				\midrule
				1   & 1.45\%                                           & [-8.41\%,  11.33\%] \\
				2   & -0.88\%                                          & [-7.87\%,  6.09\%]  \\
				3   & -0.87\%                                          & [-6.25\%,  4.51\%]  \\
				4   & -1.96\%                                          & [-6.81\%,  2.88\%]  \\
				5   & -2.19\%                                          & [-6.87\%,  2.48\%]  \\
				6   & -1.09\%                                          & [-5.23\%,  3.04\%]  \\
				7   & -0.08\%                                          & [-3.94\%,  3.78\%]  \\
				8   & 0.54\%                                           & [-2.84\%,  3.94\%]  \\
				9   & 0.85\%                                           & [-1.76\%,  3.47\%]  \\
				\bottomrule
			\end{tabular}
		\end{center}
		\caption{Calibration of cumulative inventory receives predicted $k$ weeks after ordering}
	\end{subfigure}
	\caption{Comparison of calibration metrics on-policy (control arm) versus off-policy (treatment arm) on data from real world A/B test.}
	\label{tab:genqot-offpolicy-cal}
\end{table}

\begin{table}[H]
		\begin{center}
			\begin{tabular}{ccc}
				\toprule
				         & \multicolumn{2}{c}{Change (Control - Treatment)}                        \\
				\cmidrule{2-3}
				Quantile & Mean                                             & 95\% CI              \\
				\midrule
				P10      & -0.17\%                                          & [-3.06\%,    5.49\%] \\
				P30      & 0.45\%                                           & [-4.25\%,   5.29\%]  \\
				P50      & 0.47\%                                           & [-4.53\%,    5.86\%] \\
				P70      & 0.30\%                                           & [-4.80\%,  6.88\%]   \\
				P90      & -0.56\%                                          & [-6.21\%,   7.00\%]  \\
				P98      & -6.26\%                                          & [-7.68\%,  9.61\%]   \\
				\bottomrule
			\end{tabular}
		\end{center}
	\caption{Comparison of quantile loss of VLT forecasts produced by Gen-QOT metrics on-policy (control arm) versus off-policy (treatment arm) on data from real world A/B test.}
	\label{tab:genqot-offpolicy-ql}
\end{table}

\Cref{tab:arrival-time-cal-labs} shows the difference in calibration of arrival times on the treatment and control arms of the A/B test in the actual supply chain. We see that there is not a statistically significant difference in calibration between the arms, we conclude that the degradation is not attributable to the fact that the treatment arm is ``off-policy''. For full arrival time calibration results on the A/B test data, see \Cref{sec:arrival-time-cal-results}.

\begin{figure}[h]
	\centering
	\begin{subfigure}{0.49\textwidth}
		\includegraphics[width=\textwidth]{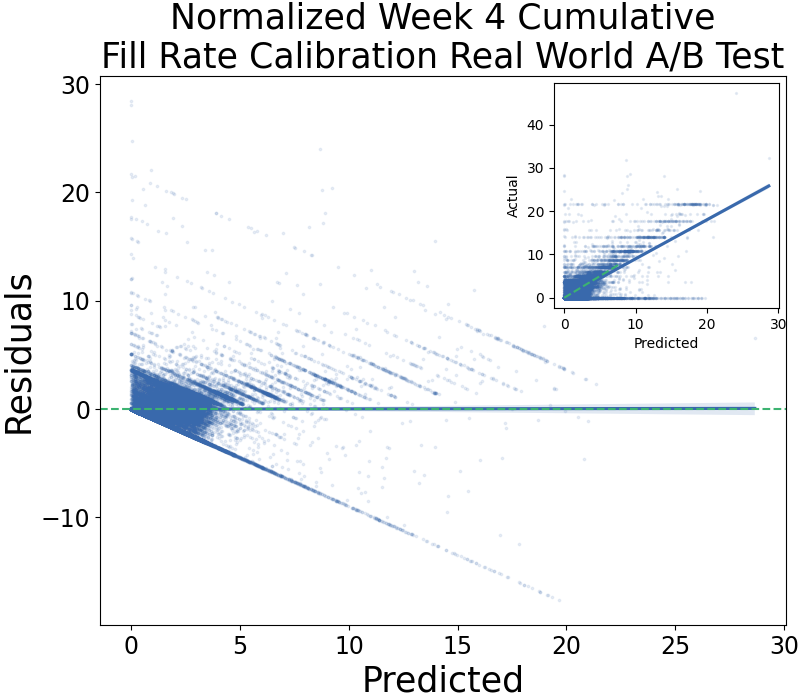}
		\caption{Cumulative receives calibration on treatment arm}
	\end{subfigure}
	\begin{subfigure}{0.49\textwidth}
		\includegraphics[width=\textwidth]{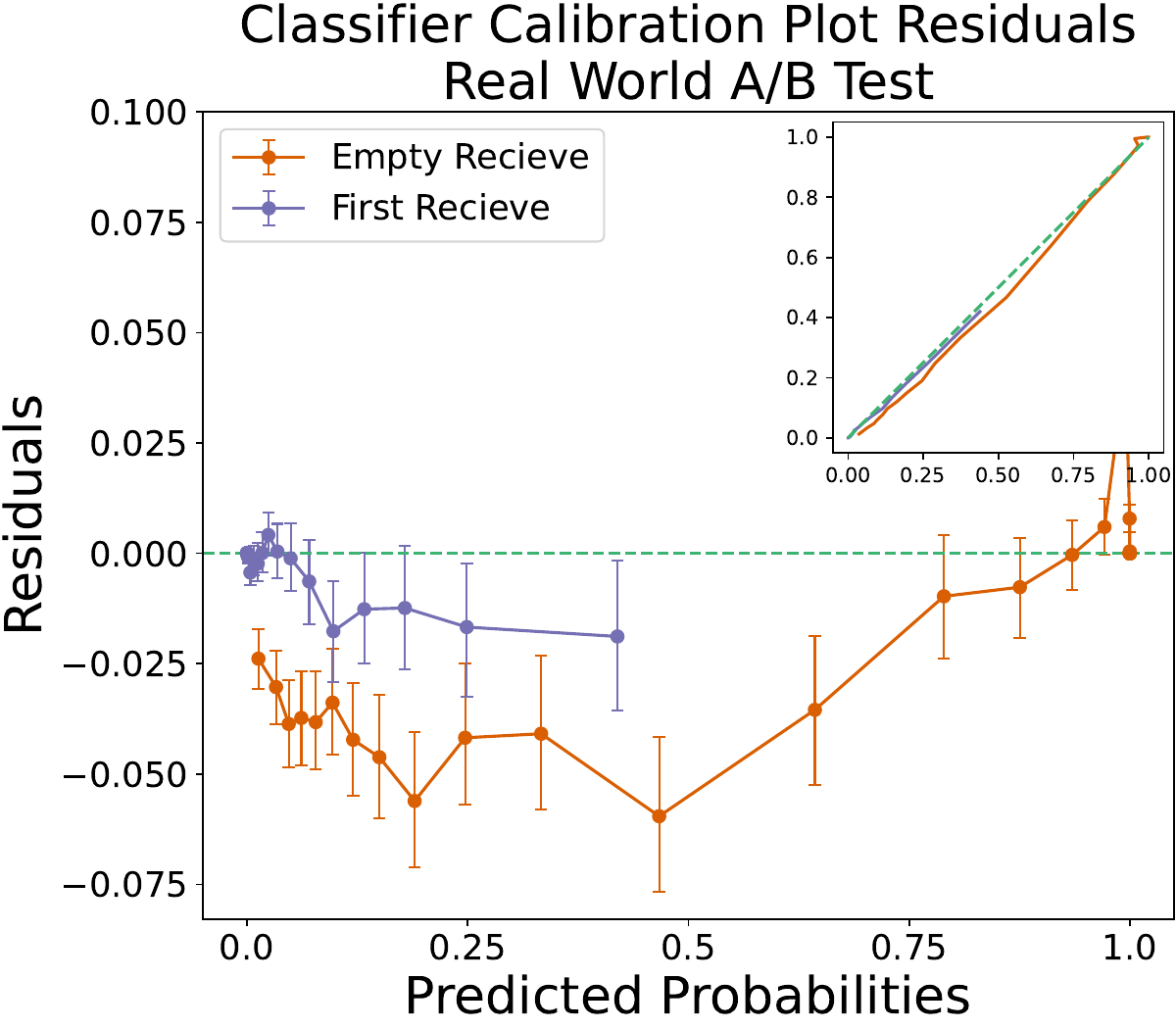}
		\caption{Classifier calibration for first receive and empty receive on treatment arm}
	\end{subfigure}
	\begin{subfigure}{0.49\textwidth}
		\includegraphics[width=\textwidth]{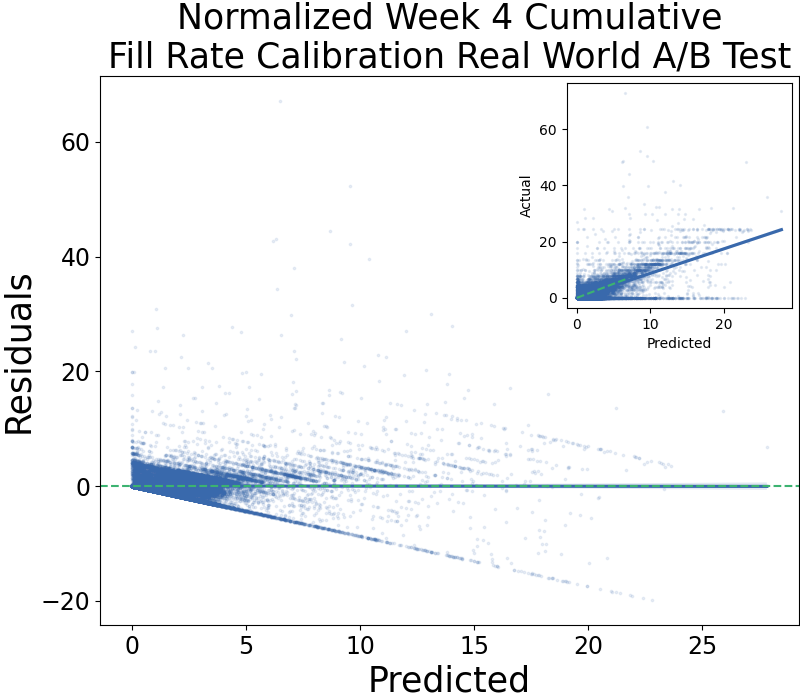}
		\caption{Cumulative receives calibration on control arm}
	\end{subfigure}
	\begin{subfigure}{0.49\textwidth}
		\includegraphics[width=\textwidth]{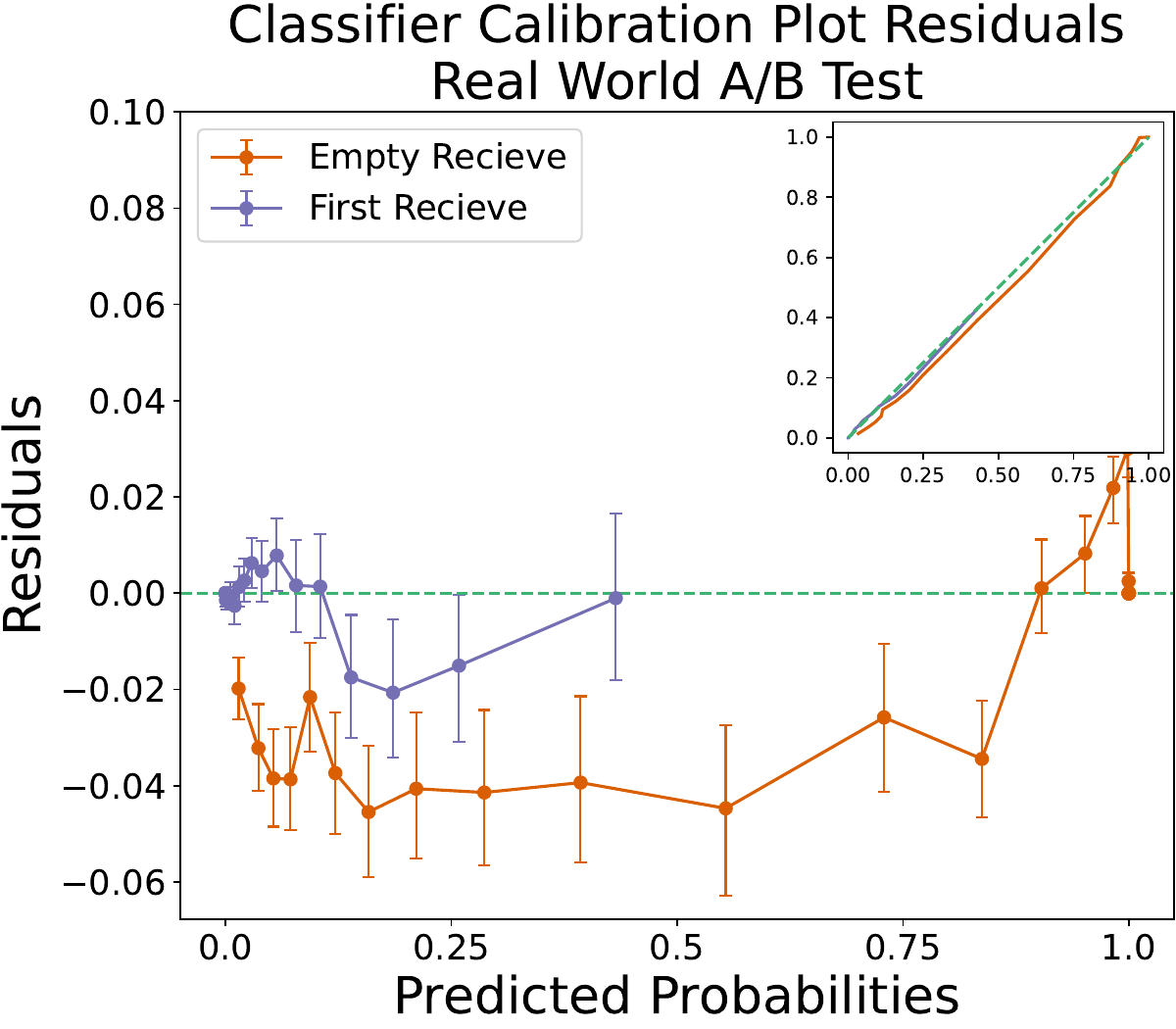}
		\caption{Classifier calibration for first receive and empty receive on control arm}
	\end{subfigure}
	\caption{Residual calibration plot for cumulative receives and classifier calibration plots on A/B test data.}
	\label{fig:labs-calibration-graphics}
\end{figure}

\Cref{fig:labs-calibration-graphics} shows Criterion 2-4 on the off-policy data. We see that the classifier calibration is still reasonable, although the cumulative receives calibration appears to have degraded from the out-of-time backtest. We emphasize that this degradation is {\it not due to the off-policy issue} as \Cref{tab:genqot-offpolicy-cal} showed that the calibration errors were statistically indistinguishable across the two arms of the A/B test. For comparison we also include the same evaluation on the control arm in \Cref{fig:labs-calibration-graphics}.

\section{Conclusion}

We extended existing work on periodic review inventory control systems to the case where inventory replenishments can arrive in multiple shipments over time. We also allow for learning an inventory control policy in the case where the retailer uses a post-processor to adjust order quantities suggested by the control policy before submitting them to the supplier -- a common practice used to ensure order quantities meet minimum order and batch sizing requirements. This is the first work to handle either of the two aforementioned complexities. We then performed extensive empirical evaluation to show the viability of the approach. We also validated that our learned dynamic model generalizes well off-policy by backtesting it on data from a real-world A/B test of our RL agent. Finally, we showed via A/B tests of the QOT-DirectBackprop agent that data-driven RL inventory control policies can outperform sophisticated base stock systems in real world settings. 

Interesting, and important, directions of future work include further exploration of the minimal assumptions needed in order for the inventory control problem to be efficiently learnable. It is also of interest to better understand the precise conditions needed on the forward dynamics model (Gen-QOT) in order for the theoretical results to hold. This may in turn give insight into what evaluations practitioners should perform on the models they incorporate into a simulator.

\clearpage
\bibliographystyle{styles/ims_nourl_eprint}
\bibliography{external}

\begin{thebibliography}{78}
\expandafter\ifx\csname natexlab\endcsname\relax\def\natexlab#1{#1}\fi
\expandafter\ifx\csname url\endcsname\relax
  \def\url#1{\texttt{#1}}\fi
\expandafter\ifx\csname urlprefix\endcsname\relax\def\urlprefix{URL }\fi

\bibitem[{Abbas et~al.(2023)Abbas, Zhao, Modayil, White and
  Machado}]{abbas2023plasticity}
\textsc{Abbas, Z.}, \textsc{Zhao, R.}, \textsc{Modayil, J.}, \textsc{White, A.}
  and \textsc{Machado, M.~C.} (2023).
\newblock Loss of plasticity in continual deep reinforcement learning.
\newblock {\href{https://arxiv.org/abs/2303.07507}{\texttt{arXiv:2303.07507}}}.

\bibitem[{Alvo et~al.(2023)Alvo, Russo and Kanoria}]{alvo2023neural}
\textsc{Alvo, M.}, \textsc{Russo, D.} and \textsc{Kanoria, Y.} (2023).
\newblock Neural inventory control in networks via hindsight differentiable
  policy optimization.
\newblock {\href{https://arxiv.org/abs/2306.11246}{\texttt{arXiv:2306.11246}}}.

\bibitem[{Arrow et~al.(1958)Arrow, Karlin, Scarf et~al.}]{arrow1958studies}
\textsc{Arrow, K.~J.}, \textsc{Karlin, S.}, \textsc{Scarf, H.~E.}
  \textsc{et~al.} (1958).
\newblock \textit{Studies in the mathematical theory of inventory and
  production}.
\newblock Stanford University Press.

\bibitem[{Augenblick and Rabin(2019)}]{augenblick2018belief}
\textsc{Augenblick, N.} and \textsc{Rabin, M.} (2019).
\newblock {Belief movement, uncertainty reduction, and rational updating}.
\newblock Tech. rep., Haas School of Business, University of California,
  Berkeley.

\bibitem[{Balaji et~al.(2019)Balaji, Bell-Masterson, Bilgin, Damianou, Garcia,
  Jain, Luo, Maggiar, Narayanaswamy and Ye}]{balaji2019orl}
\textsc{Balaji, B.}, \textsc{Bell-Masterson, J.}, \textsc{Bilgin, E.},
  \textsc{Damianou, A.}, \textsc{Garcia, P.~M.}, \textsc{Jain, A.},
  \textsc{Luo, R.}, \textsc{Maggiar, A.}, \textsc{Narayanaswamy, B.} and
  \textsc{Ye, C.} (2019).
\newblock Orl: Reinforcement learning benchmarks for online stochastic
  optimization problems.
\newblock {\href{https://arxiv.org/abs/1911.10641}{\texttt{arXiv:1911.10641}}}.

\bibitem[{Bao(2006)}]{bao2006supply}
\textsc{Bao, Y.} (2006).
\newblock Supply chain competition.
\newblock Tech. rep., UNSW Sydney.
{\href{https://unsworks.unsw.edu.au/bitstreams/5031b434-336e-4a5f-8fde-0f97ff29af55/download}{[PDF]}}

\bibitem[{Bollapragada and Morton(1999)}]{bollapragada1999myopic}
\textsc{Bollapragada, S.} and \textsc{Morton, T.~E.} (1999).
\newblock Myopic heuristics for the random yield problem.
\newblock \textit{Operations Research} \textbf{47} 713--722.

\bibitem[{Brown et~al.(2020)Brown, Mann, Ryder, Subbiah, Kaplan, Dhariwal,
  Neelakantan, Shyam, Sastry, Askell, Agarwal, Herbert-Voss, Krueger, Henighan,
  Child, Ramesh, Ziegler, Wu, Winter, Hesse, Chen, Sigler, Litwin, Gray, Chess,
  Clark, Berner, McCandlish, Radford, Sutskever and Amodei}]{brown2020language}
\textsc{Brown, T.}, \textsc{Mann, B.}, \textsc{Ryder, N.}, \textsc{Subbiah,
  M.}, \textsc{Kaplan, J.~D.}, \textsc{Dhariwal, P.}, \textsc{Neelakantan, A.},
  \textsc{Shyam, P.}, \textsc{Sastry, G.}, \textsc{Askell, A.},
  \textsc{Agarwal, S.}, \textsc{Herbert-Voss, A.}, \textsc{Krueger, G.},
  \textsc{Henighan, T.}, \textsc{Child, R.}, \textsc{Ramesh, A.},
  \textsc{Ziegler, D.}, \textsc{Wu, J.}, \textsc{Winter, C.}, \textsc{Hesse,
  C.}, \textsc{Chen, M.}, \textsc{Sigler, E.}, \textsc{Litwin, M.},
  \textsc{Gray, S.}, \textsc{Chess, B.}, \textsc{Clark, J.}, \textsc{Berner,
  C.}, \textsc{McCandlish, S.}, \textsc{Radford, A.}, \textsc{Sutskever, I.}
  and \textsc{Amodei, D.} (2020).
\newblock Language models are few-shot learners.
\newblock In \textit{Advances in Neural Information Processing Systems}
  (H.~Larochelle, M.~Ranzato, R.~Hadsell, M.~Balcan and H.~Lin, eds.), vol.~33.
  Curran Associates, Inc.

\bibitem[{Chan and Muckstadt(1999)}]{chan1999effects}
\textsc{Chan, E.} and \textsc{Muckstadt, J.} (1999).
\newblock The effects of load smoothing on inventory levels in a capacitated
  production and inventory system.
\newblock Tech. rep., Cornell University Operations Research and Industrial
  Engineering.
{\href{https://ecommons.cornell.edu/bitstreams/027de6ee-6247-4358-997b-c66eaedc3bf8/download}{[PDF]}}

\bibitem[{Clavera et~al.(2020)Clavera, Fu and Abbeel}]{clavera2020model}
\textsc{Clavera, I.}, \textsc{Fu, V.} and \textsc{Abbeel, P.} (2020).
\newblock Model-augmented actor-critic: Backpropagating through paths.
\newblock In \textit{ICLR}.

\bibitem[{Dada et~al.(2007)Dada, Petruzzi and Schwarz}]{dada2007news}
\textsc{Dada, M.}, \textsc{Petruzzi, N.~C.} and \textsc{Schwarz, L.~B.} (2007).
\newblock A newsvendor’s procurement problem when suppliers are unreliable.
\newblock \textit{Manufacturing \& Service Operations Management} \textbf{9}
  9--32.

\bibitem[{Das et~al.(1999)Das, Gosavi, Mahadevan and
  Marchalleck}]{das1999solving}
\textsc{Das, T.~K.}, \textsc{Gosavi, A.}, \textsc{Mahadevan, S.} and
  \textsc{Marchalleck, N.} (1999).
\newblock Solving semi-markov decision problems using average reward
  reinforcement learning.
\newblock \textit{Management Science} \textbf{45} 560--574.

\bibitem[{Dawid(1982)}]{dawid1982well}
\textsc{Dawid, A.} (1982).
\newblock The well calibrated bayesian.
\newblock \textit{Journal of the American Statistical Association} \textbf{77}
  605--613.

\bibitem[{Devlin et~al.(2019)Devlin, Chang, Lee and Toutanova}]{devlin2019bert}
\textsc{Devlin, J.}, \textsc{Chang, M.-W.}, \textsc{Lee, K.} and
  \textsc{Toutanova, K.} (2019).
\newblock {BERT: Pre-training of Deep Bidirectional Transformers for Language
  Understanding}.
\newblock In \textit{NAACL-HLT}.

\bibitem[{Efroni et~al.(2022{\natexlab{a}})Efroni, Foster, Misra, Krishnamurthy
  and Langford}]{efroni2022sample}
\textsc{Efroni, Y.}, \textsc{Foster, D.~J.}, \textsc{Misra, D.},
  \textsc{Krishnamurthy, A.} and \textsc{Langford, J.} (2022{\natexlab{a}}).
\newblock Sample-efficient reinforcement learning in the presence of exogenous
  information.
\newblock {\href{https://arxiv.org/abs/2206.04282}{\texttt{arXiv:2206.04282}}}.

\bibitem[{Efroni et~al.(2022{\natexlab{b}})Efroni, Kakade, Krishnamurthy and
  Zhang}]{efroni2022sparsity}
\textsc{Efroni, Y.}, \textsc{Kakade, S.}, \textsc{Krishnamurthy, A.} and
  \textsc{Zhang, C.} (2022{\natexlab{b}}).
\newblock Sparsity in partially controllable linear systems.
\newblock In \textit{International Conference on Machine Learning}. PMLR.

\bibitem[{Eisenach et~al.(2020)Eisenach, Patel and Madeka}]{eisenach2020mqt}
\textsc{Eisenach, C.}, \textsc{Patel, Y.} and \textsc{Madeka, D.} (2020).
\newblock {MQTransformer: Multi-Horizon Forecasts with Context Dependent and
  Feedback-Aware Attention}.
\newblock {\href{https://arxiv.org/abs/2009.14799}{\texttt{arXiv:2009.14799}}}.

\bibitem[{Federgruen and Zipkin(1986)}]{federgruen1986inventory}
\textsc{Federgruen, A.} and \textsc{Zipkin, P.} (1986).
\newblock An inventory model with limited production capacity and uncertain
  demands i. the average-cost criterion.
\newblock \textit{Mathematics of Operations Research} \textbf{11} 193--207.

\bibitem[{Fisher and Raman(1996)}]{fisher1996reducing}
\textsc{Fisher, M.} and \textsc{Raman, A.} (1996).
\newblock Reducing the cost of demand uncertainty through accurate response to
  early sales.
\newblock \textit{Operations research} \textbf{44} 87--99.

\bibitem[{Foster and Stine(2021)}]{foster2021threshold}
\textsc{Foster, D.} and \textsc{Stine, R.} (2021).
\newblock {Threshold Martingales and the Evolution of Forecasts}.
\newblock {\href{https://arxiv.org/abs/2105.06834}{\texttt{arXiv:2105.06834}}}.

\bibitem[{Foster and Vohra(1997)}]{foster1997calibrated}
\textsc{Foster, D.~P.} and \textsc{Vohra, R.~V.} (1997).
\newblock Calibrated learning and correlated equilibrium.
\newblock \textit{Games and Economic Behavior} \textbf{21} 40--55.

\bibitem[{Gasparin et~al.(2019)Gasparin, Lukovic and Alippi}]{gasparin2019deep}
\textsc{Gasparin, A.}, \textsc{Lukovic, S.} and \textsc{Alippi, C.} (2019).
\newblock {Deep Learning for Time Series Forecasting: The Electric Load Case}.
\newblock {\href{https://arxiv.org/abs/1907.09207}{\texttt{arXiv:1907.09207}}}.

\bibitem[{Gerchak et~al.(1988)Gerchak, Vickson and
  Parlar}]{gerchak1988periodic}
\textsc{Gerchak, Y.}, \textsc{Vickson, R.~G.} and \textsc{Parlar, M.} (1988).
\newblock Periodic review production models with variable yield and uncertain
  demand.
\newblock \textit{Iie Transactions} \textbf{20} 144--150.

\bibitem[{Giannoccaro and Pontrandolfo(2002)}]{giannoccaro2002inventory}
\textsc{Giannoccaro, I.} and \textsc{Pontrandolfo, P.} (2002).
\newblock Inventory management in supply chains: a reinforcement learning
  approach.
\newblock \textit{International Journal of Production Economics} \textbf{78}
  153--161.

\bibitem[{Gijsbrechts et~al.(2022)Gijsbrechts, Boute, Van~Mieghem and
  Zhang}]{gijsbrechts2022can}
\textsc{Gijsbrechts, J.}, \textsc{Boute, R.~N.}, \textsc{Van~Mieghem, J.~A.}
  and \textsc{Zhang, D.~J.} (2022).
\newblock Can deep reinforcement learning improve inventory management?
  performance on lost sales, dual-sourcing, and multi-echelon problems.
\newblock \textit{Manufacturing \& Service Operations Management} \textbf{24}
  1349--1368.

\bibitem[{Goodfellow et~al.(2014)Goodfellow, Pouget-Abadie, Mirza, Xu,
  Warde-Farley, Ozair, Courville and Bengio}]{goodfellow2014gan}
\textsc{Goodfellow, I.}, \textsc{Pouget-Abadie, J.}, \textsc{Mirza, M.},
  \textsc{Xu, B.}, \textsc{Warde-Farley, D.}, \textsc{Ozair, S.},
  \textsc{Courville, A.} and \textsc{Bengio, Y.} (2014).
\newblock Generative adversarial nets.
\newblock In \textit{Advances in Neural Information Processing Systems}
  (Z.~Ghahramani, M.~Welling, C.~Cortes, N.~Lawrence and K.~Weinberger, eds.),
  vol.~27. Curran Associates, Inc.

\bibitem[{Graves(2012)}]{graves2012sequence}
\textsc{Graves, A.} (2012).
\newblock Sequence transduction with recurrent neural networks.
\newblock {\href{https://arxiv.org/abs/1211.3711}{\texttt{arXiv:1211.3711}}}.

\bibitem[{Graves(2013)}]{graves2013generating}
\textsc{Graves, A.} (2013).
\newblock Generating sequences with recurrent neural networks.
\newblock {\href{https://arxiv.org/abs/1308.0850}{\texttt{arXiv:1308.0850}}}.

\bibitem[{Henig and Gerchak(1990)}]{henig1990structure}
\textsc{Henig, M.} and \textsc{Gerchak, Y.} (1990).
\newblock The structure of periodic review policies in the presence of random
  yield.
\newblock \textit{Operations Research} \textbf{38} 634--643.

\bibitem[{Hu et~al.(2019)Hu, Liu, Spielberg, Tenenbaum, Freeman, Wu, Rus and
  Matusik}]{hu2019chainqueen}
\textsc{Hu, Y.}, \textsc{Liu, J.}, \textsc{Spielberg, A.}, \textsc{Tenenbaum,
  J.~B.}, \textsc{Freeman, W.~T.}, \textsc{Wu, J.}, \textsc{Rus, D.} and
  \textsc{Matusik, W.} (2019).
\newblock Chainqueen: A real-time differentiable physical simulator for soft
  robotics.
\newblock In \textit{2019 International conference on robotics and automation
  (ICRA)}. IEEE.

\bibitem[{Ingraham et~al.(2018)Ingraham, Riesselman, Sander and
  Marks}]{ingraham2018learning}
\textsc{Ingraham, J.}, \textsc{Riesselman, A.}, \textsc{Sander, C.} and
  \textsc{Marks, D.} (2018).
\newblock Learning protein structure with a differentiable simulator.
\newblock In \textit{International Conference on Learning Representations}.

\bibitem[{Januschowski et~al.(2022)Januschowski, Wang, Torkkola, Erkkilä,
  Hasson and Gasthaus}]{januschowski2022forecasting}
\textsc{Januschowski, T.}, \textsc{Wang, Y.}, \textsc{Torkkola, K.},
  \textsc{Erkkilä, T.}, \textsc{Hasson, H.} and \textsc{Gasthaus, J.} (2022).
\newblock Forecasting with trees.
\newblock \textit{International Journal of Forecasting} \textbf{38} 1473--1481.
\newblock Special Issue: M5 competition.

\bibitem[{Kaplan(1970)}]{kaplan1970dynamic}
\textsc{Kaplan, R.~S.} (1970).
\newblock A dynamic inventory model with stochastic lead times.
\newblock \textit{Management Science} \textbf{16} 491--507.

\bibitem[{Kiesm{\"u}ller et~al.(2011)Kiesm{\"u}ller, Kok and
  Dabia}]{kiesmller2011single}
\textsc{Kiesm{\"u}ller, G.~P.}, \textsc{Kok, D.} and \textsc{Dabia, S.} (2011).
\newblock Single item inventory control under periodic review and a minimum
  order quantity.
\newblock \textit{International Journal of Production Economics} \textbf{133}
  280--285.

\bibitem[{Kingma and Welling(2013)}]{kingma2013auto}
\textsc{Kingma, D.~P.} and \textsc{Welling, M.} (2013).
\newblock Auto-encoding variational bayes.
\newblock {\href{https://arxiv.org/abs/1312.6114}{\texttt{arXiv:1312.6114}}}.

\bibitem[{Kingma and Welling(2019)}]{kingma2019intro}
\textsc{Kingma, D.~P.} and \textsc{Welling, M.} (2019).
\newblock An introduction to variational autoencoders.
\newblock {\href{https://arxiv.org/abs/1906.02691}{\texttt{arXiv:1906.02691}}}.

\bibitem[{Li et~al.(2004)Li, Xu and Hayya}]{li2004periodic}
\textsc{Li, Z.}, \textsc{Xu, S.~H.} and \textsc{Hayya, J.} (2004).
\newblock A periodic-review inventory system with supply interruptions.
\newblock \textit{Probability in the Engineering and Informational Sciences}
  \textbf{18} 33–53.

\bibitem[{Lim et~al.(2019)Lim, Arik, Loeff and Pfister}]{lim2019tft}
\textsc{Lim, B.}, \textsc{Arik, S.~O.}, \textsc{Loeff, N.} and \textsc{Pfister,
  T.} (2019).
\newblock {Temporal Fusion Transformers for Interpretable Multi-horizon Time
  Series Forecasting}.
\newblock {\href{https://arxiv.org/abs/1912.09363}{\texttt{arXiv:1912.09363}}}.

\bibitem[{Maddah and Jaber(2008)}]{maddah2008economic}
\textsc{Maddah, B.} and \textsc{Jaber, M.~Y.} (2008).
\newblock Economic order quantity for items with imperfect quality: Revisited.
\newblock \textit{International Journal of Production Economics} \textbf{112}
  808--815.
\newblock Special Section on RFID: Technology, Applications, and Impact on
  Business Operations.

\bibitem[{Madeka et~al.(2018)Madeka, Swiniarski, Foster, Razoumov, Torkkola and
  Wen}]{madeka2018sample}
\textsc{Madeka, D.}, \textsc{Swiniarski, L.}, \textsc{Foster, D.},
  \textsc{Razoumov, L.}, \textsc{Torkkola, K.} and \textsc{Wen, R.} (2018).
\newblock Sample path generation for probabilistic demand forecasting.
\newblock In \textit{KDD 2018 Workshop on Mining and Learning from Time
  Series}.

\bibitem[{Madeka et~al.(2022)Madeka, Torkkola, Eisenach, Luo, Foster and
  Kakade}]{madeka2022deep}
\textsc{Madeka, D.}, \textsc{Torkkola, K.}, \textsc{Eisenach, C.}, \textsc{Luo,
  A.}, \textsc{Foster, D.} and \textsc{Kakade, S.} (2022).
\newblock Deep inventory management.
\newblock {\href{https://arxiv.org/abs/2210.03137}{\texttt{arXiv:2210.03137}}}.

\bibitem[{Maggiar et~al.(2022)Maggiar, Song and
  Muharremoglu}]{maggiar2022multi}
\textsc{Maggiar, A.}, \textsc{Song, I.} and \textsc{Muharremoglu, A.} (2022).
\newblock Multi-echelon inventory management for a non-stationary capacitated
  distribution network.
\newblock Tech. rep., SSRN.
{\href{https://papers.ssrn.com/sol3/Delivery.cfm/SSRN_ID4154780_code1172651.pdf?abstractid=4154780&mirid=1&type=2}{[PDF]}}

\bibitem[{Mnih et~al.(2016)Mnih, Badia, Mirza, Graves, Lillicrap, Harley,
  Silver and Kavukcuoglu}]{mnih2016asynchronous}
\textsc{Mnih, V.}, \textsc{Badia, A.~P.}, \textsc{Mirza, M.}, \textsc{Graves,
  A.}, \textsc{Lillicrap, T.~P.}, \textsc{Harley, T.}, \textsc{Silver, D.} and
  \textsc{Kavukcuoglu, K.} (2016).
\newblock Asynchronous methods for deep reinforcement learning.
\newblock {\href{https://arxiv.org/abs/1602.01783}{\texttt{arXiv:1602.01783}}}.

\bibitem[{Mnih et~al.(2013)Mnih, Kavukcuoglu, Silver, Graves, Antonoglou,
  Wierstra and Riedmiller}]{mnih2013playing}
\textsc{Mnih, V.}, \textsc{Kavukcuoglu, K.}, \textsc{Silver, D.},
  \textsc{Graves, A.}, \textsc{Antonoglou, I.}, \textsc{Wierstra, D.} and
  \textsc{Riedmiller, M.} (2013).
\newblock Playing atari with deep reinforcement learning.
\newblock {\href{https://arxiv.org/abs/1312.5602}{\texttt{arXiv:1312.5602}}}.

\bibitem[{Mousa et~al.(2023)Mousa, van~de Berg, Kotecha, del Rio-Chanona and
  Mowbray}]{mousa2023analysis}
\textsc{Mousa, M.}, \textsc{van~de Berg, D.}, \textsc{Kotecha, N.}, \textsc{del
  Rio-Chanona, E.~A.} and \textsc{Mowbray, M.} (2023).
\newblock An analysis of multi-agent reinforcement learning for decentralized
  inventory control systems.
\newblock {\href{https://arxiv.org/abs/2307.11432}{\texttt{arXiv:2307.11432}}}.

\bibitem[{Mukhoty et~al.(2019)Mukhoty, Maurya and Shukla}]{mukhoty2019seq}
\textsc{Mukhoty, B.~P.}, \textsc{Maurya, V.} and \textsc{Shukla, S.~K.} (2019).
\newblock {Sequence to sequence deep learning models for solar irradiation
  forecasting}.
\newblock In \textit{IEEE Milan PowerTech}.

\bibitem[{Nahmias(1979)}]{nahmias1979simple}
\textsc{Nahmias, S.} (1979).
\newblock Simple approximations for a variety of dynamic leadtime lost-sales
  inventory models.
\newblock \textit{Operations Research} \textbf{27} 904--924.

\bibitem[{Nascimento et~al.(2019)Nascimento, Souto, Ogasawara, Porto and
  Bezerra}]{nascimento2019stconv}
\textsc{Nascimento, R.~C.}, \textsc{Souto, Y.~M.}, \textsc{Ogasawara, E.},
  \textsc{Porto, F.} and \textsc{Bezerra, E.} (2019).
\newblock {STConvS2S: Spatiotemporal Convolutional Sequence to Sequence Network
  for weather forecasting}.
\newblock {\href{https://arxiv.org/abs/1912.00134}{\texttt{arXiv:1912.00134}}}.

\bibitem[{Parmas et~al.(2023)Parmas, Seno and Aoki}]{parmas2023model}
\textsc{Parmas, P.}, \textsc{Seno, T.} and \textsc{Aoki, Y.} (2023).
\newblock Model-based reinforcement learning with scalable composite policy
  gradient estimators.
\newblock In \textit{ICML}.

\bibitem[{Porteus(2002)}]{porteus2002foundations}
\textsc{Porteus, E.~L.} (2002).
\newblock \textit{Foundations of stochastic inventory theory}.
\newblock Stanford University Press.

\bibitem[{Qi et~al.(2023)Qi, Shi, Qi, Ma, Yuan, Wu and Shen}]{qi2023practical}
\textsc{Qi, M.}, \textsc{Shi, Y.}, \textsc{Qi, Y.}, \textsc{Ma, C.},
  \textsc{Yuan, R.}, \textsc{Wu, D.} and \textsc{Shen, Z.-J.} (2023).
\newblock A practical end-to-end inventory management model with deep learning.
\newblock \textit{Management Science} \textbf{69} 759--773.

\bibitem[{Robb and Silver(1998)}]{robb1998inventory}
\textsc{Robb, D.~J.} and \textsc{Silver, E.~A.} (1998).
\newblock Inventory management with periodic ordering and minimum order
  quantities.
\newblock \textit{Journal of the Operational Research Society} \textbf{49}
  1085--1094.

\bibitem[{Salinas et~al.(2020)Salinas, Flunkert, Gasthaus and
  Januschowski}]{salinas2020deepar}
\textsc{Salinas, D.}, \textsc{Flunkert, V.}, \textsc{Gasthaus, J.} and
  \textsc{Januschowski, T.} (2020).
\newblock Deepar: Probabilistic forecasting with autoregressive recurrent
  networks.
\newblock \textit{International Journal of Forecasting} \textbf{36} 1181--1191.

\bibitem[{Schulman et~al.(2017)Schulman, Wolski, Dhariwal, Radford and
  Klimov}]{schulman2017proximal}
\textsc{Schulman, J.}, \textsc{Wolski, F.}, \textsc{Dhariwal, P.},
  \textsc{Radford, A.} and \textsc{Klimov, O.} (2017).
\newblock Proximal policy optimization algorithms.
\newblock {\href{https://arxiv.org/abs/1707.06347}{\texttt{arXiv:1707.06347}}}.

\bibitem[{Shen et~al.(2019)Shen, Tian and Zhu}]{shen2019two}
\textsc{Shen, H.}, \textsc{Tian, T.} and \textsc{Zhu, H.} (2019).
\newblock A two-echelon inventory system with a minimum order quantity
  requirement.
\newblock \textit{Sustainability} \textbf{11}.

\bibitem[{Silver et~al.(2016)Silver, Huang, Maddison, Guez, Sifre, Van
  Den~Driessche, Schrittwieser, Antonoglou, Panneershelvam, Lanctot
  et~al.}]{silver2016mastering}
\textsc{Silver, D.}, \textsc{Huang, A.}, \textsc{Maddison, C.~J.},
  \textsc{Guez, A.}, \textsc{Sifre, L.}, \textsc{Van Den~Driessche, G.},
  \textsc{Schrittwieser, J.}, \textsc{Antonoglou, I.}, \textsc{Panneershelvam,
  V.}, \textsc{Lanctot, M.} \textsc{et~al.} (2016).
\newblock Mastering the game of go with deep neural networks and tree search.
\newblock \textit{Nature} \textbf{529} 484--489.

\bibitem[{Sinclair et~al.(2023)Sinclair, Vieira~Frujeri, Cheng, Marshall,
  Barbalho, Li, Neville, Menache and Swaminathan}]{sinclair2023hindsight}
\textsc{Sinclair, S.~R.}, \textsc{Vieira~Frujeri, F.}, \textsc{Cheng, C.-A.},
  \textsc{Marshall, L.}, \textsc{Barbalho, H. D.~O.}, \textsc{Li, J.},
  \textsc{Neville, J.}, \textsc{Menache, I.} and \textsc{Swaminathan, A.}
  (2023).
\newblock Hindsight learning for {MDP}s with exogenous inputs.
\newblock In \textit{Proceedings of the 40th International Conference on
  Machine Learning}, vol. 202 of \textit{Proceedings of Machine Learning
  Research}. PMLR.

\bibitem[{Song and Zipkin(1996)}]{song1996inventory}
\textsc{Song, J.-S.} and \textsc{Zipkin, P.~H.} (1996).
\newblock Inventory control with information about supply conditions.
\newblock \textit{Management Science} \textbf{42} 1409--1419.

\bibitem[{Suh et~al.(2022)Suh, Simchowitz, Zhang and
  Tedrake}]{suh2022differentiable}
\textsc{Suh, H.~J.}, \textsc{Simchowitz, M.}, \textsc{Zhang, K.} and
  \textsc{Tedrake, R.} (2022).
\newblock Do differentiable simulators give better policy gradients?
\newblock In \textit{International Conference on Machine Learning}. PMLR.

\bibitem[{Sundermeyer et~al.(2010)Sundermeyer, Schluter and
  Ney}]{sundermeyer2010lstm}
\textsc{Sundermeyer, M.}, \textsc{Schluter, R.} and \textsc{Ney, H.} (2010).
\newblock {LSTM Neural Networks for Language Modeling}.
\newblock In \textit{INTERSPEECH}.

\bibitem[{Sutton and Barto(2020)}]{sutton2020reinforcement}
\textsc{Sutton, R.~S.} and \textsc{Barto, A.~G.} (2020).
\newblock \textit{Reinforcement Learning: An iIntroduction}.
\newblock MIT press.

\bibitem[{Szepesvári(2010)}]{szepesvari2010algorithms}
\textsc{Szepesvári, C.} (2010).
\newblock \textit{Algorithms for Reinforcement Learning}.
\newblock Synthesis Lectures on Artificial Intelligence and Machine Learning,
  Morgan \& Claypool Publishers.

\bibitem[{Taleb(2018)}]{taleb2018election}
\textsc{Taleb, N.~N.} (2018).
\newblock Election predictions as martingales: an arbitrage approach.
\newblock \textit{Quantitative Finance} \textbf{18} 1--5.

\bibitem[{Taleb and Madeka(2019)}]{taleb2019all}
\textsc{Taleb, N.~N.} and \textsc{Madeka, D.} (2019).
\newblock All roads lead to quantitative finance.
\newblock \textit{Quantitative Finance} \textbf{19} 1775--1776.

\bibitem[{THOMAS(2023)}]{thomas2023towards}
\textsc{THOMAS, J.~D.} (2023).
\newblock Towards cooperative marl in industrial domains .

\bibitem[{van~den Oord et~al.(2016)van~den Oord, Dieleman, Zen, Simonyan,
  Vinyals, Graves, Kalchbrenner, Senior and Kavukcuoglu}]{oord2016wavenet}
\textsc{van~den Oord, A.}, \textsc{Dieleman, S.}, \textsc{Zen, H.},
  \textsc{Simonyan, K.}, \textsc{Vinyals, O.}, \textsc{Graves, A.},
  \textsc{Kalchbrenner, N.}, \textsc{Senior, A.} and \textsc{Kavukcuoglu, K.}
  (2016).
\newblock Wavenet: A generative model for raw audio.
\newblock {\href{https://arxiv.org/abs/1609.03499}{\texttt{arXiv:1609.03499}}}.

\bibitem[{Van~Oord et~al.(2016)Van~Oord, Kalchbrenner and
  Kavukcuoglu}]{van2016pixel}
\textsc{Van~Oord, A.}, \textsc{Kalchbrenner, N.} and \textsc{Kavukcuoglu, K.}
  (2016).
\newblock Pixel recurrent neural networks.
\newblock In \textit{International conference on machine learning}. PMLR.

\bibitem[{Veinott(1965)}]{veinott1965optimal}
\textsc{Veinott, A.~F.} (1965).
\newblock The optimal inventory policy for batch ordering.
\newblock \textit{Operations Research} \textbf{13} 424--432.

\bibitem[{Wen and Torkkola(2019)}]{wen2019deep}
\textsc{Wen, R.} and \textsc{Torkkola, K.} (2019).
\newblock {Deep Generative Quantile-Copula Models for Probabilistic
  Forecasting}.
\newblock In \textit{ICML Time Series Workshop}.

\bibitem[{Wen et~al.(2017)Wen, Torkkola, Narayanaswamy and
  Madeka}]{wen2017mqcnn}
\textsc{Wen, R.}, \textsc{Torkkola, K.}, \textsc{Narayanaswamy, B.} and
  \textsc{Madeka, D.} (2017).
\newblock {A multi-horizon quantile recurrent forecaster}.
\newblock In \textit{NIPS Time Series Workshop}.

\bibitem[{Williams and Zipser(1989)}]{Williams1989}
\textsc{Williams, R.~J.} and \textsc{Zipser, D.} (1989).
\newblock A learning algorithm for continually running fully recurrent neural
  networks.
\newblock \textit{Neural Computation} \textbf{1} 270--280.

\bibitem[{Yu et~al.(2017)Yu, Zheng, Anandkumar and Yue}]{yu2017long}
\textsc{Yu, R.}, \textsc{Zheng, S.}, \textsc{Anandkumar, A.} and \textsc{Yue,
  Y.} (2017).
\newblock {Long-term Forecasting using Higher Order Tensor RNNs}.
\newblock {\href{https://arxiv.org/abs/1711.00073}{\texttt{arXiv:1711.00073}}}.

\bibitem[{Zhao et~al.(2023)Zhao, Tang and Yao}]{zhao2023policy}
\textsc{Zhao, H.}, \textsc{Tang, W.} and \textsc{Yao, D.~D.} (2023).
\newblock Policy optimization for continuous reinforcement learning.
\newblock {\href{https://arxiv.org/abs/2305.18901}{\texttt{arXiv:2305.18901}}}.

\bibitem[{Zhao and Katehakis(2006)}]{zhao2006structure}
\textsc{Zhao, Y.} and \textsc{Katehakis, M.~N.} (2006).
\newblock On the structure of optimal ordering policies for stochastic
  inventory systems with minimum order quantity.
\newblock \textit{Probability in the Engineering and Informational Sciences}
  \textbf{20} 257–270.

\bibitem[{Zhou et~al.(2007)Zhou, Zhao and Katehakis}]{zhou2007effective}
\textsc{Zhou, B.}, \textsc{Zhao, Y.} and \textsc{Katehakis, M.~N.} (2007).
\newblock Effective control policies for stochastic inventory systems with a
  minimum order quantity and linear costs.
\newblock \textit{International Journal of Production Economics} \textbf{106}
  523--531.

\bibitem[{Zhu(2022)}]{zhu2022simple}
\textsc{Zhu, H.} (2022).
\newblock A simple heuristic policy for stochastic inventory systems with both
  minimum and maximum order quantity requirements.
\newblock \textit{Annals of Operations Research} \textbf{309} 347–363.

\bibitem[{Zhu et~al.(2015)Zhu, Liu and Chen}]{zhu2015effective}
\textsc{Zhu, H.}, \textsc{Liu, X.} and \textsc{Chen, Y.~F.} (2015).
\newblock Effective inventory control policies with a minimum order quantity
  and batch ordering.
\newblock \textit{International Journal of Production Economics} \textbf{168}
  21--30.

\bibitem[{Zipkin(2008)}]{zipkin2008old}
\textsc{Zipkin, P.} (2008).
\newblock Old and new methods for lost-sales inventory systems.
\newblock \textit{Operations research} \textbf{56} 1256--1263.

\end{thebibliography}

\appendix
\clearpage

\section{Gen-QOT Metrics}
\label{sec:metrics}
In this section we define the metrics used to evaluate Gen-QOT.

\subsection{Vendor Lead Time Forecasting Metrics}
For the purposes of defining the vendor lead time metric, assume we have a set of historic purchase orders. Each sample $i \in \cS$ consist of a sequence of arrivals $\{(\kappa^i_1, l^i_1),\dots,(\kappa^i_{J}, l^i_{J}) \}$ where $j$ denotes the sequence index with some maximum sequence length $J$, $\kappa^i_j$ denotes the arrival quantity, and $l^i_j$ denotes the lead-time of the arrival relative to some forecast creation time, which can be the order date or any day after.

The quantile loss of a forecast at quantile $q$ is defined as $\mathrm{QL}_q(x,y) = (1-q)(x-y)^+ + q(y-x)^+$. For a fixed quantile $q$, the quantile loss metric evaluated in this paper is defined as
\begin{equation}
	\label{eqn:QL-lt}
	\mathrm{QL}_q := \frac{\sum_{i \in \mathcal{S} } \sum_{j=1}^{J} \kappa^i_{j} \mathrm{QL}_{q}(l^i_{j}, \hat{l}^{i, q})}{ \sum_{i \in \mathcal{S} }  \sum_{j=1}^{J} \kappa^i_{j}}
\end{equation}

Next, to compute the CRPS, we assume access to a quantile estimate of the lead-time distribution for each sample, $\hat{l}^{i, q}$ over a set of quantiles $\cQ := \{0.01,...,0.99\}$. The  CRPS is computed approximately by averaging over the quantile losses for each quantile in $q \in \cQ$.

\begin{equation}
	\label{eqn:crps-lt}
	\mathrm{CRPS} := \frac{\sum_{i \in \mathcal{S} } \sum_{j=1}^{J} \kappa^i_{j} \frac{1}{|\cQ|} \sum_{q\in\cQ} \mathrm{QL}_{q}(l^i_{j}, \hat{l}^{i, q})
	}{ \sum_{i \in \mathcal{S} }  \sum_{j=1}^{J} \kappa^i_{j}}
\end{equation}

\subsection{Calibration Metrics}
\label{sec:metrics-cal}
A forecast of the probabilities of a sequence of events $E_1,E_2,...$ where $E_t \in \{0,1\}$ -- is said to be calibrated if whenever a forecast $p$ of $E_t=1$ is made, the empirical probabilities are $\approx p$. Because the probability is real valued, the interval $[0,1]$ is split into bins in order to get empirical probabilities whenever the forecast was $p$.

In the case of estimating the mean of a random variable (such as percent of order received after $l$ weeks), we define calibration as the regression coefficient of a simple linear regression of the actual value given the predicted value. If a forecast is well calibrated, this coefficient should be 1.

See \citet{foster1997calibrated,dawid1982well} for a more in-depth discussion of calibrated forecasts.

\subsection{Arrival Time Calibration}
\label{sec:arrival-time-cal}
Here we assess if the forecast probability of receiving all the inventory from an order by a specific date is well calibrated according to the Gen-QOT model.

We treat this problem as a classification task by assigning a class label of one to all periods before, and including, the period of the terminal arrival and zero to all periods after. This indicates whether a final receive for an order occurs by a specific date. Using the samples drawn from Gen-QOT, we estimate the probability of a final receive by a date, and evaluate the calibration of the predicted distributions using both the predicted probabilities and actual class labels. Bucketing receive predictions into deciles, we measure the mean class label and receive probability for each bucket. If our model is well calibrated, we expect the mean predicted probability to fall within the confidence interval for the expected actual label.

\clearpage
\section{Additional Numerical Results for Gen-QOT}

\subsection{Arrival Time Calibration -- Full Results}
\label{sec:arrival-time-cal-results}

In \Cref{tab:cal-arrival-in-time-full}, \Cref{tab:cal-arrival-out-time-full}, \Cref{tab:cal-arrival-ctrl-full}, and \Cref{tab:cal-arrival-trt-full} we present the same arrival time calibration as \Cref{sec:eval-dynamics}, but now split out by lead time. For lead time, probability bin pairs with less than 10 samples we omit the results.

\begin{table}[H]
	\caption{Calibration of probabilistic arrival time forecasts for lead times $l=1$ to $l=8$ on an in-of-time holdout set}
	\label{tab:cal-arrival-in-time-full}
	\begin{center}
		\begin{tabular}{ccccc}
			\toprule
			Probability & $l=1$         & $l=2$         & $l=3$         & $l=4$         \\
			\midrule
			0.0-0.1     & 0.080 ± 0.003 & 0.066 ± 0.001 & 0.068 ± 0.001 & 0.057 ± 0.000 \\
			0.1-0.2     & 0.126 ± 0.003 & 0.164 ± 0.002 & 0.160 ± 0.001 & 0.151 ± 0.001 \\
			0.2-0.3     & 0.301 ± 0.006 & 0.252 ± 0.002 & 0.242 ± 0.001 & 0.247 ± 0.001 \\
			0.3-0.4     & 0.414 ± 0.007 & 0.374 ± 0.002 & 0.358 ± 0.002 & 0.339 ± 0.002 \\
			0.4-0.5     & 0.540 ± 0.006 & 0.471 ± 0.002 & 0.443 ± 0.002 & 0.419 ± 0.002 \\
			0.5-0.6     & 0.606 ± 0.006 & 0.564 ± 0.001 & 0.535 ± 0.002 & 0.525 ± 0.002 \\
			0.6-0.7     & 0.710 ± 0.004 & 0.663 ± 0.001 & 0.631 ± 0.002 & 0.597 ± 0.003 \\
			0.7-0.8     & 0.790 ± 0.003 & 0.745 ± 0.001 & 0.740 ± 0.002 & 0.691 ± 0.004 \\
			0.8-0.9     & 0.882 ± 0.001 & 0.840 ± 0.001 & 0.813 ± 0.002 & 0.830 ± 0.004 \\
			0.9-1.0     & 0.984 ± 0.000 & 0.936 ± 0.001 & 0.923 ± 0.002 & 0.911 ± 0.003 \\
			\cmidrule{2-5}
			            & $l=5$         & $l=6$         & $l=7$         & $l=8$         \\
			\cmidrule{2-5}
			0.0-0.1     & 0.050 ± 0.000 & 0.045 ± 0.000 & 0.041 ± 0.000 & 0.037 ± 0.000 \\
			0.1-0.2     & 0.147 ± 0.001 & 0.142 ± 0.001 & 0.130 ± 0.001 & 0.140 ± 0.001 \\
			0.2-0.3     & 0.237 ± 0.001 & 0.212 ± 0.001 & 0.222 ± 0.002 & 0.228 ± 0.002 \\
			0.3-0.4     & 0.315 ± 0.002 & 0.304 ± 0.002 & 0.302 ± 0.003 & 0.291 ± 0.003 \\
			0.4-0.5     & 0.392 ± 0.003 & 0.386 ± 0.003 & 0.348 ± 0.004 & 0.346 ± 0.005 \\
			0.5-0.6     & 0.474 ± 0.003 & 0.438 ± 0.004 & 0.503 ± 0.005 & 0.521 ± 0.007 \\
			0.6-0.7     & 0.566 ± 0.004 & 0.633 ± 0.006 & 0.514 ± 0.007 & 0.655 ± 0.018 \\
			0.7-0.8     & 0.744 ± 0.005 & 0.644 ± 0.006 & 0.689 ± 0.015 & -- \\
			0.8-0.9     & 0.787 ± 0.005 & 0.802 ± 0.007 & 0.750 ± 0.045 & -- \\
			0.9-1.0     & 0.880 ± 0.005 & -- & -- & -- \\
			\bottomrule
		\end{tabular}
	\end{center}
\end{table}

\begin{table}[H]
	\caption{Calibration of probabilistic arrival time forecasts for lead times $l=1$ to $l=8$ on an out-of-time holdout set}
	\label{tab:cal-arrival-out-time-full}
	\begin{center}
		\begin{tabular}{ccccc}
			\toprule
			Probability & $l=1$         & $l=2$         & $l=3$         & $l=4$         \\
			\midrule
			0.0-0.1     & 0.072 ± 0.002 & 0.059 ± 0.001 & 0.065 ± 0.000 & 0.056 ± 0.000 \\
			0.1-0.2     & 0.163 ± 0.003 & 0.159 ± 0.001 & 0.154 ± 0.001 & 0.152 ± 0.001 \\
			0.2-0.3     & 0.285 ± 0.004 & 0.262 ± 0.001 & 0.250 ± 0.001 & 0.245 ± 0.001 \\
			0.3-0.4     & 0.409 ± 0.004 & 0.357 ± 0.001 & 0.346 ± 0.001 & 0.335 ± 0.001 \\
			0.4-0.5     & 0.485 ± 0.004 & 0.468 ± 0.001 & 0.438 ± 0.001 & 0.423 ± 0.001 \\
			0.5-0.6     & 0.612 ± 0.004 & 0.564 ± 0.001 & 0.540 ± 0.001 & 0.498 ± 0.002 \\
			0.6-0.7     & 0.693 ± 0.003 & 0.645 ± 0.001 & 0.644 ± 0.001 & 0.596 ± 0.002 \\
			0.7-0.8     & 0.793 ± 0.002 & 0.756 ± 0.001 & 0.723 ± 0.001 & 0.680 ± 0.003 \\
			0.8-0.9     & 0.887 ± 0.001 & 0.843 ± 0.001 & 0.815 ± 0.002 & 0.809 ± 0.003 \\
			0.9-1.0     & 0.983 ± 0.000 & 0.932 ± 0.001 & 0.922 ± 0.001 & 0.907 ± 0.002 \\
			\cmidrule{2-5}
			            & $l=5$         & $l=6$         & $l=7$         & $l=8$         \\
			\cmidrule{2-5}
			0.0-0.1     & 0.050 ± 0.000 & 0.046 ± 0.000 & 0.040 ± 0.000 & 0.038 ± 0.000 \\
			0.1-0.2     & 0.144 ± 0.001 & 0.136 ± 0.001 & 0.126 ± 0.001 & 0.134 ± 0.001 \\
			0.2-0.3     & 0.230 ± 0.001 & 0.218 ± 0.001 & 0.213 ± 0.001 & 0.203 ± 0.001 \\
			0.3-0.4     & 0.321 ± 0.001 & 0.302 ± 0.001 & 0.281 ± 0.002 & 0.254 ± 0.002 \\
			0.4-0.5     & 0.398 ± 0.002 & 0.364 ± 0.002 & 0.344 ± 0.003 & 0.351 ± 0.003 \\
			0.5-0.6     & 0.461 ± 0.002 & 0.461 ± 0.003 & 0.428 ± 0.003 & 0.516 ± 0.005 \\
			0.6-0.7     & 0.559 ± 0.003 & 0.550 ± 0.003 & 0.548 ± 0.005 & 0.662 ± 0.013 \\
			0.7-0.8     & 0.710 ± 0.003 & 0.666 ± 0.004 & 0.612 ± 0.010 & 0.739 ± 0.023 \\
			0.8-0.9     & 0.781 ± 0.003 & 0.724 ± 0.005 & 0.893 ± 0.004 & 1.000 ± 0.000 \\
			0.9-1.0     & 0.863 ± 0.004 & 0.963 ± 0.001 & 1.000 ± 0.000 & 1.000 ± 0.000 \\
			\bottomrule
		\end{tabular}
	\end{center}
\end{table}

\begin{table}[H]
	\caption{Calibration of probabilistic arrival time forecasts for lead times $l=1$ to $l=8$ on control arm of real-world A/B test}
	\label{tab:cal-arrival-ctrl-full}
	\begin{center}
		\begin{tabular}{ccccc}
			\toprule
			Probability & $l=1$         & $l=2$         & $l=3$         & $l=4$         \\
			\midrule
			0.0-0.1     & 0.175 ± 0.011 & 0.164 ± 0.006 & 0.166 ± 0.003 & 0.155 ± 0.001 \\
			0.1-0.2     & 0.216 ± 0.008 & 0.303 ± 0.006 & 0.302 ± 0.002 & 0.302 ± 0.002 \\
			0.2-0.3     & 0.483 ± 0.013 & 0.367 ± 0.005 & 0.417 ± 0.002 & 0.402 ± 0.002 \\
			0.3-0.4     & 0.548 ± 0.012 & 0.471 ± 0.004 & 0.510 ± 0.002 & 0.492 ± 0.002 \\
			0.4-0.5     & 0.712 ± 0.011 & 0.601 ± 0.003 & 0.581 ± 0.002 & 0.586 ± 0.002 \\
			0.5-0.6     & 0.642 ± 0.010 & 0.672 ± 0.002 & 0.669 ± 0.002 & 0.657 ± 0.002 \\
			0.6-0.7     & 0.733 ± 0.009 & 0.755 ± 0.002 & 0.720 ± 0.002 & 0.711 ± 0.002 \\
			0.7-0.8     & 0.841 ± 0.005 & 0.815 ± 0.001 & 0.801 ± 0.002 & 0.811 ± 0.002 \\
			0.8-0.9     & 0.897 ± 0.002 & 0.885 ± 0.001 & 0.870 ± 0.002 & 0.890 ± 0.002 \\
			0.9-1.0     & 0.988 ± 0.000 & 0.959 ± 0.000 & 0.950 ± 0.001 & 0.953 ± 0.001 \\
			\cmidrule{2-5}
			            & $l=5$         & $l=6$         & $l=7$         & $l=8$         \\
			\cmidrule{2-5}
			0.0-0.1     & 0.138 ± 0.001 & 0.126 ± 0.001 & 0.110 ± 0.001 & 0.088 ± 0.000 \\
			0.1-0.2     & 0.281 ± 0.001 & 0.279 ± 0.001 & 0.264 ± 0.001 & 0.228 ± 0.001 \\
			0.2-0.3     & 0.394 ± 0.002 & 0.382 ± 0.002 & 0.377 ± 0.002 & 0.338 ± 0.002 \\
			0.3-0.4     & 0.498 ± 0.002 & 0.487 ± 0.003 & 0.467 ± 0.003 & 0.397 ± 0.002 \\
			0.4-0.5     & 0.566 ± 0.003 & 0.572 ± 0.003 & 0.580 ± 0.003 & 0.500 ± 0.003 \\
			0.5-0.6     & 0.649 ± 0.003 & 0.646 ± 0.003 & 0.637 ± 0.003 & 0.582 ± 0.004 \\
			0.6-0.7     & 0.716 ± 0.003 & 0.755 ± 0.003 & 0.720 ± 0.003 & 0.589 ± 0.009 \\
			0.7-0.8     & 0.823 ± 0.002 & 0.802 ± 0.002 & 0.763 ± 0.004 & -- \\
			0.8-0.9     & 0.887 ± 0.002 & 0.894 ± 0.002 & --  & -- \\
			0.9-1.0     & 0.935 ± 0.001 & 0.842 ± 0.006 & -- & -- \\
			\bottomrule
		\end{tabular}
	\end{center}
\end{table}

\begin{table}[H]
	\caption{Calibration of probabilistic arrival time forecasts for lead times $l=1$ to $l=8$ on treatment arm of real-world A/B test}
	\label{tab:cal-arrival-trt-full}
	\begin{center}
		\begin{tabular}{ccccc}
			\toprule
			Probability & $l=1$         & $l=2$         & $l=3$         & $l=4$         \\
			\midrule
			0.0-0.1     & 0.137 ± 0.005 & 0.123 ± 0.004 & 0.147 ± 0.002 & 0.141 ± 0.001 \\
			0.1-0.2     & 0.388 ± 0.010 & 0.295 ± 0.006 & 0.291 ± 0.002 & 0.277 ± 0.001 \\
			0.2-0.3     & 0.453 ± 0.013 & 0.344 ± 0.004 & 0.410 ± 0.002 & 0.382 ± 0.002 \\
			0.3-0.4     & 0.483 ± 0.013 & 0.481 ± 0.003 & 0.485 ± 0.002 & 0.484 ± 0.002 \\
			0.4-0.5     & 0.687 ± 0.010 & 0.597 ± 0.002 & 0.575 ± 0.002 & 0.562 ± 0.002 \\
			0.5-0.6     & 0.741 ± 0.007 & 0.668 ± 0.002 & 0.648 ± 0.002 & 0.659 ± 0.002 \\
			0.6-0.7     & 0.740 ± 0.007 & 0.752 ± 0.002 & 0.729 ± 0.002 & 0.725 ± 0.002 \\
			0.7-0.8     & 0.826 ± 0.004 & 0.813 ± 0.001 & 0.812 ± 0.002 & 0.801 ± 0.002 \\
			0.8-0.9     & 0.908 ± 0.002 & 0.887 ± 0.001 & 0.884 ± 0.001 & 0.887 ± 0.002 \\
			0.9-1.0     & 0.988 ± 0.000 & 0.958 ± 0.000 & 0.951 ± 0.001 & 0.950 ± 0.001 \\
			\cmidrule{2-5}
			            & $l=5$         & $l=6$         & $l=7$         & $l=8$         \\
			\cmidrule{2-5}
			0.0-0.1     & 0.126 ± 0.001 & 0.113 ± 0.001 & 0.097 ± 0.001 & 0.078 ± 0.000 \\
			0.1-0.2     & 0.260 ± 0.001 & 0.263 ± 0.001 & 0.254 ± 0.001 & 0.229 ± 0.001 \\
			0.2-0.3     & 0.371 ± 0.002 & 0.377 ± 0.002 & 0.370 ± 0.002 & 0.329 ± 0.002 \\
			0.3-0.4     & 0.481 ± 0.002 & 0.476 ± 0.003 & 0.442 ± 0.003 & 0.392 ± 0.002 \\
			0.4-0.5     & 0.564 ± 0.003 & 0.537 ± 0.003 & 0.515 ± 0.003 & 0.474 ± 0.003 \\
			0.5-0.6     & 0.624 ± 0.003 & 0.617 ± 0.003 & 0.618 ± 0.003 & 0.576 ± 0.004 \\
			0.6-0.7     & 0.726 ± 0.003 & 0.704 ± 0.003 & 0.684 ± 0.003 & 0.707 ± 0.008 \\
			0.7-0.8     & 0.796 ± 0.003 & 0.813 ± 0.002 & 0.749 ± 0.003 & -- \\
			0.8-0.9     & 0.869 ± 0.002 & 0.863 ± 0.002 & 0.833 ± 0.017 & -- \\
			0.9-1.0     & 0.930 ± 0.001 & 0.852 ± 0.005 & --			& -- \\
			\bottomrule
		\end{tabular}
	\end{center}
\end{table}

\subsection{Neural Architecture Ablation}
\label{sec:neural-ablation}

Given the broad set of neural architectures available for fitting {\it sequence-to-sequence} problems, we test a set of different encoder and decoder neural networks classes. Specifically we tested multi-layer perceptron vs causal-convolution encoder, and recurrent neural network vs transformer decoder. In the end, we trained four models on data from sequences of arrivals from 10MM  orders from 2017 and 2018 and tested on arrivals from  50K orders from 2019. To assess prediction quality, we rely on negative-log-likelihood of next token prediction, as well as unit weighted quantile-loss of cumulative quantity arrivals at 1, 4, and 9 weeks since order was placed.

\begin{table}[H]
	\caption{Results of ablation analysis for multiple metrics.}
	\begin{center}
		\begin{tabular}{lccccc}
			\toprule
			Model           & \multicolumn{5}{c}{Metric}                                                                                                                           \\
			\midrule
			                & \multicolumn{5}{c}{QL of Cumulative Quantity of Arrivals: Week1\vspace{0.25em}}                                                                      \\

			                & P10                                                                             & P30            & P50            & P70            & P90             \\
			\cmidrule{2-6}
			MLP-RNN         & 100.00                                                                          & 100.00         & 100.00         & 100.00         & \textbf{100.00} \\
			CNN-RNN         & 70.15                                                                           & 98.60          & \textbf{89.15} & \textbf{96.67} & 100.42          \\
			CNN-Transformer & \textbf{68.54}                                                                  & \textbf{95.79} & 91.76          & 98.81          & 110.10          \\
			\midrule
			                & \multicolumn{5}{c}{QL of Cumulative Quantity of Arrivals: Week4\vspace{0.25em}}                                                                      \\
			                & P10                                                                             & P30            & P50            & P70            & P90             \\
			\cmidrule{2-6}
			MLP-RNN         & 100.00                                                                          & 100.00         & 100.00         & 100.00         & 100.00          \\
			CNN-RNN         & 89.43                                                                           & 95.90          & 97.48          & 100.00         & 104.94          \\
			CNN-Transformer & \textbf{29.27}                                                                  & \textbf{39.85} & \textbf{52.04} & \textbf{66.75} & \textbf{80.86}  \\
			\midrule
			                & \multicolumn{5}{c}{QL of Cumulative Quantity of Arrivals: Week9\vspace{0.25em}}                                                                      \\
			                & P10                                                                             & P30            & P50            & P70            & P90             \\
			\cmidrule{2-6}
			MLP-RNN         & 100.00                                                                          & 100.00         & 100.00         & 100.00         & \textbf{100.00} \\
			CNN-RNN         & 100.79                                                                          & 100.58         & 100.82         & 101.70         & 102.01          \\
			CNN-Transformer & \textbf{90.24}                                                                  & \textbf{92.24} & \textbf{96.29} & \textbf{99.72} & 104.03          \\
			\midrule
			                & \multicolumn{5}{c}{Negative Log-Likelihood of Next Token Prediction}                                                                                 \\
			\cmidrule{2-6}
			MLP-RNN         & \multicolumn{5}{c}{100.00}                                                                                                                           \\
			CNN-RNN         & \multicolumn{5}{c}{\textbf{93.76}}                                                                                                                   \\
			CNN-Transformer & \multicolumn{5}{c}{94.07}                                                                                                                            \\
			\bottomrule
		\end{tabular}
		\label{table:qot-ablation-results}
	\end{center}
\end{table}

\clearpage
\section{Featurization}
\label{sec:featurization}
\subsubsection*{Buying Agent}
Below are the features provided to the RL policy -- they are the same as \citet{madeka2022deep}. Specifically, the state at time $t$ for product $i$ contains:
\begin{enumerate}
	\item The current inventory level $I^i_{(t-1)}$
	\item Previous actions $a^i_u$ that have been taken for all $u < t$
	\item Demand time series features
	      \begin{enumerate}
		      \item Historical availability corrected demand
		      \item Distance to public holidays
		      \item Historical website glance views data
	      \end{enumerate}
	\item Static product features
	      \begin{enumerate}
		      \item Product group
		      \item Text-based features from the product description
		      \item Brand
	      \end{enumerate}
	\item Economics of the product - (price, cost etc.)
\end{enumerate}

\subsubsection*{Gen-QOT Model}
The exogenous context and other information provided to Gen-QOT at time $t$ for product $i$ contains:
\begin{enumerate}
	\item Current action $a_t^i$
	\item Previous actions $a^i_u$ that have been taken for all $u < t$
	\item Time series features
	      \begin{enumerate}
		      \item Distance to public holidays
		      \item Previous arrivals for all times $u < t$
		      \item Vendor constraints (minimum order quantities, batch sizes, etc.)
	      \end{enumerate}
	\item Static product features
	      \begin{enumerate}
		      \item Product group
		      \item Brand
		      \item Vendor
	      \end{enumerate}
\end{enumerate}

\section{The Gen-QOT Model}
\label{sec:genqot}
In this section, we describe our novel arrivals prediction model. First, denote historical covariates $x^i_t$ for each product $i$ at each time $t$ that will be used to estimate the distribution of state transition probabilities. The vector $x^i_t$ can be thought of as the observational historical data that contains at minimum, information such as the time-series of historical orders and arrivals. Other information that can be incorporated includes vendor and product attributes, existing geographic inventory allocation, and the distance to various holidays that may affect the ability for vendors and logistics providers to reliably fulfill and ship inventory. See \Cref{sec:featurization} for the list of features we used. Also denote the actual arrivals of inventory as $\{o^i_{t,j}\}_{j=0}^L$, where $o^i_{t,j} \in \mathbb{R}_{\geq 0}$. Recall $L$ is the maximum possible lead time for an arrival.

The Gen-QOT model solves the problem of predicting the joint distribution of inventory arrivals for an action $a^i_t$ at time $t$. To model the distribution of arrival sequences $\{o^i_{t,j}\}_{j=1}^L$, we consider the distribution over partial fill rates\footnote{These are the same as the $\bar{\alpha}_{t,j}^i$ in \Cref{sec:method-sim}.} instead:
\[
	\alpha^i_{t,j} := \begin{cases}
		\frac{o^i_{t,j}}{a^i_t} & a^i_t \neq 0 \\
		0                       & a^i_t = 0.   \\
	\end{cases}
\]
Our goal then is to produce a generative model from which we can sample sequences of partial fill rates $\{\alpha^i_{t,j}\}_{j=1}^L$. Note that these partial fill rates {\it do not} need to sum to 1  as this is modeling the ``end-to-end'' yield (see \Cref{fig:fills-system}).

Next, observe that an equivalent formulation is to model a sequence of tuples of partial fill rates and time since the last non-zero arrival: $\{ (k^i_{t,s}, \tilde{\alpha}^i_{t,s})\}_{s \in \mathbb{Z}_{\geq 0}}$, where $\tilde{\alpha}^i_{t,s}$ denotes the proportion of $a^i_t$ in the $s^{th}$ arrival and $k^i_{t,s}$ denotes the number of periods since the previous non-zero arrival. By convention the first arrival is measured as an offset from $t-1$.

\subsection{Probabilistic Model}

At a high-level, the methodology we employ to predict a sequence of arrivals is to construct a grid over quantity and time that can be used to bin individual arrivals into distinct arrival classes. Our model then produce a categorical distribution over these classes conditioned on previous arrivals in the sequence, akin to generative sequence modeling in NLP. Once the sequence of classes has been sampled, we can map it back to the original quantity of interest using the function $V$, defined in \Cref{sec:generating-samples}.

\subsubsection{Model the distribution by binning}

We rely on binning to avoid making parametric assumptions about the distributions of arrivals. To model the proportion of requested inventory in each arrival and number weeks since last arrival, we can bin these pairs into classes. We define a sequence of $N$ grid points  $\tau^{(1)},\dots,\tau^{(N)}$  over time periods since last arrival and $M$ grid points $p^{(1)},\dots, p^{(M)}$ for the proportions, such that
\begin{align*}
	 & \tau^{(n)} \leq \tau^{(m)}      ~~\forall n < m, \\
	 & p^{(n)} \leq p^{(m)}            ~~\forall n < m, \\
	 & \tau^{(1)} = p^{(1)} = 0,                        \\
	 & \tau^{(N)} = \tau_{max}                          \\
	 & p^{(M)} = p_{max}
\end{align*}
where $\tau_{max}$ and $p_{max}$ are both large constants. For each $n,m \in [N-1]\times[M-1]$, define the set
\[
	Z_{n,m} := \{ (k,\alpha) : \tau^{(n)} \leq k < \tau^{(n+1)} \text{ and } p^{(m)} \leq \alpha < p^{(m+1)} \}.
\]
By construction, $\mathcal{Z} := \{ Z_{n,m} : (n,m) \in [N-1]\times[M-1] \, \}$ is a partition of $\{u \in \mathbb{Z} \, | \, 0 \leq u < \tau_{max} \;\} \times \{v \in \mathbb{R} \, | \, 0 \leq v < p_{max} \;\}$, the space of all possible pairs $(k^i_{s}, \tilde{\alpha}^i_{t,s})$.

Next we denote the index pair $(n,m)$ that corresponds to a specific arrival class $Z_{n,m}$ by the random vector $\zeta$. For example $\zeta = (1,2)$ is a reference to the arrival class $Z_{1,2}$. Additionally, we add a special index pair denoting the end of an arrivals sequence, which we can signify with $\cancel{\zeta}$. This $\cancel{\zeta}$ can be thought of as a special arrival that designates that the sequence of arrivals from an action has terminated. In practice, we can use $(0,0)$ to represent the value of $\cancel{\zeta}$. Given this construction, we have $\zeta$ taking values in $\tilde{\mathcal{Z}} := \{[N-1]\times[M-1])\cup \{(0,0)\}\}$

Our objective is now to estimate the joint probability of a sequence of index pairs $\zeta^i_{t,0}, \zeta^i_{t,1} \dots$ corresponding to a sequence of arrival classes given some $a^i_t$. Gen-QOT specifically estimates
\begin{equation}
	P(\zeta^i_{t,0}, \zeta^i_{t,1}, ... | x^i_{t}, a^i_{t,0}) =
	\prod_{j} P(\zeta^i_{t,j}| \zeta^i_{t,j-1}, \zeta^i_{t,j-2}, \dots, \zeta^i_{t,0}, x^i_{t}, a^i_{t})
\end{equation}
and decomposes this joint probability into the product of conditional probabilities.

\subsubsection{Generating samples of inventory arrivals: mapping from classes to actual arrivals}
\label{sec:generating-samples}

Sampling this joint probability distribution gives a sequence of arrival classes $(\zeta^i_{t,0}, \zeta^i_{t,1}, \dots )$. To obtain a sample path of inventory arrivals for an action $a^i_t$, we use the function $V: \tilde{\mathcal{Z}} \rightarrow \mathbb{R}_{> 0} \times \mathbb{Z}_{> 0} $ to map each element $\zeta^i_{t,s} = (n,m)$ of the sampled sequences of arrival classes back to a tuple of partial fill rates and time-since-last-arrival. This function is implemented by mapping each of the tuples $(n,m)$ to a representative element $(\bar{k}_{n,m}$, $\bar{\alpha}_{n,m})$ in the corresponding $Z_{i,j}$. Here, $\bar{k}_{n,m}$ and $\bar{\alpha}_{n,m}$ denote the mean of the respective quantities observed in the data in each arrival class bin, for every $n,m$. However, we note that other quantities can be used, e.g. the center of the bin of each arrival class $(\frac{\tau^{(n)} + \tau^{(n+1)}}{2}, \frac{p^{(i)} + p^{(i+1)}}{2})$.

The sampled sequence of arrival class indexes, $(\zeta^i_{t,0}, \zeta^i_{t,1}, \dots )$, can be transformed back into the original quantity of interest $o^i_{t,j}$ by substituting each $\zeta^i_{t,s}$ with the tuple $(\bar{k}_{n,m},  \bar{\alpha}_{n,m})$ to recover a sequence of estimated periods since last arrival and proportion of inventory, $(k^i_{t,s}, \tilde{\alpha}^i_{t,s})$. By cumulatively summing the periods, and multiplying each $\tilde{\alpha}^i_{t,s}$ by $a^i_t$, we immediately recover every $o^i_{t,j}$.

\subsubsection{Example of arrival sequence transformation}

\textbf{Inventory arrivals to arrival class sequence:} As an example, let $a^i_t=10$, and the true arrival sequence be $\langle0,3,5,0,4\rangle$. Additionally, we can construct a series of grid-points $\tau^{(l)} = l - 1$ and $p^{(m)}=0.2\cdot (m - 1)$  for all $l \in [L]$ and $m \in [M]$ where $L=4$ and $M=6$.

Then we can map from the original sequence of arrivals over time $\{o^i_{t,j}\}$ to tuples of partial fill rates and time since last arrival $\{ (k^i_{s}, \tilde{\alpha}^i_{t,s})\}$ by normalizing by action and computing the time since last arrival.
\[
	\langle0,3,5,0,4\rangle ~\longrightarrow~ \langle(2,0.3),(1,0.5),(2,0.4)\rangle
\]
Using the grid-points and $(0,0)$ as a reference for $\cancel{\zeta}$, we can transform the sequence of $\{ (k^i_{s}, \tilde{\alpha}^i_{t,s})\}$ into a sequence of $\zeta$'s by binning the tuples of time since last arrival and inventory percentile. For example the tuple $(2,0.3)$ is placed in the bin with coordinates $(2,2)$ because it is $2$ time periods since the start of the sample and $0.3$ falls in the second bin that sits between $0.2$ and $0.4$. The rest of the sequence is transformed as follows:
\[
	\langle(2,0.3),(1,0.5),(2,0.4)\rangle ~\longrightarrow~ \langle(2,2),(1,3),(2,3),(0,0)\rangle
\]
\textbf{Arrival class sequence to inventory arrivals:} We can reverse this example by imagining Gen-QOT generated the sequence $\langle(2,2),(1,3),(2,3),(0,0)\rangle$. We can replace each $Z_{n,m}$ with the representative element corresponding to the middle of the bin $(\bar{k}_{n,m},  \bar{\alpha}_{n,m}) = (l, 0.2 \cdot m - 0.1)$. Substituting into the sequence we get
\[
	\langle(2,2),(1,3),(2,3),(0,0)\rangle ~\longrightarrow~ \langle(2,0.3),(1,0.5),(2,0.5)\rangle
\]
Then we can cumulatively sum the time since last arrival and multiply by the action to recover sampled sequence of arrivals over time, $\{o^i_{t,j}\}$
\[
	\langle(2,0.3),(1,0.5),(2,0.5)\rangle ~\longrightarrow~ \langle0,3,5,0,5\rangle
\]

In this example we see two crucial features of the probabilistic model utilized by Gen-QOT. Firstly, the sum of arrival quantities over actual and sampled arrival quantities do not need to be equal to the action $a^i_t$. Secondly, the structure of the grid and choice of representative unit can induce error in the estimated sample if not carefully chosen.

\subsection{Neural Architecture and Loss}
\label{sec:nn}
Following canonical work in generative modeling for language \citep{sundermeyer2010lstm,graves2013generating}, our work uses recurrent neural-networks to generate sequences of arrival classes. Additionally, following \citet{oord2016wavenet,wen2017mqcnn} we rely on stacked and dilated temporal convolutions to learn a useful embedding of historical time series data. Merging the architectures together, Gen-QOT is implemented using a encoder-decoder style architecture with a dilated temporal convolution encoder and recurrent decoder. Full model hyperparameters along with arrival class definitions can be found in \Cref{sec:hyperparameters}. Additionally, a richer comparison of various model architectures can be found in  \Cref{sec:neural-ablation}.

\begin{figure}[!htb]
	\centering
	\includegraphics[width=0.8\columnwidth]{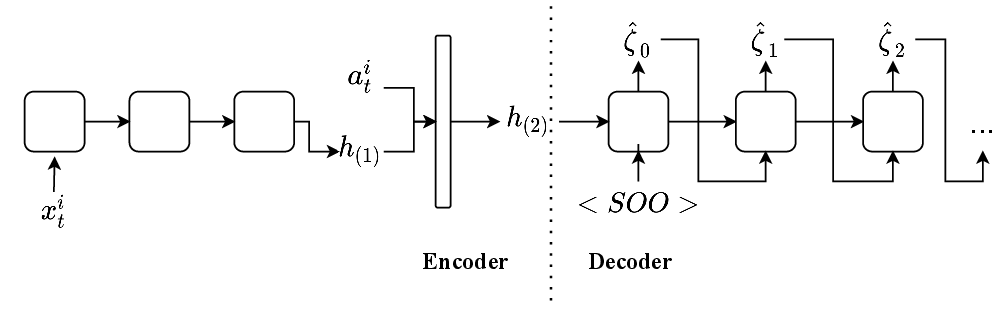}
	\caption[width=5cm]{Visualization of Gen-QOT model architecture. The model structure uses a classic encoder decoder architecture with a dilated causal convolution encoder and standard recurrent decoder. The diagram above demonstrates how samples are generated the Gen-QOT model during inference, where $<SOO>$ is a vector of zeros}
	\label{fig:rl_qot_model}
\end{figure}

The network is optimized to maximize the likelihood of generated samples by being trained to minimize cross-entropy loss. Allowing $y$ to be a matrix of indicators across all arrival classes $Z_{n,m}$ and $\hat{y}$ to be the matrix of probabilities for each arrival class, then the loss for a single prediction can be written as
\[
	J(y, \hat{y}\,) =  - \sum^{N-1}_{n=1} \sum^{M-1}_{m=1}  y_{n,m} \log(\,\hat{y}_{n,m}\,)
\]
where the sum runs over all arrival classes. Finally, the network is trained using the teacher-forcing algorithm \citep{Williams1989}, where during training the model learns to predict the next token in a sequence given the actual trailing sub-sequence. During inference, strategies like beam-search \citep{graves2012sequence} can be used to find the highest likelihood sequence of arrival classes. For our work, we implemented a simplified inference algorithm that samples the predicted distribution of arrival classes to generate a trailing sub-sequence that is used to predict subsequent token.

\subsection{Gen-QOT Training Hyperparameters}
\label{sec:hyperparameters}

\begin{table}[H]
	\caption{Training hyperparameters for Gen-QOT}
	\begin{center}
		\begin{tabular}{l c}
			\toprule
			Hyper-parameter                & Value            \\
			\midrule
			Epochs                         & $500$            \\
			Learning Rate                  & $1\times10^{-4}$ \\
			Optimizer                      & Adam             \\
			Number of Convolution Layers   & $5$              \\
			Number of Convolution Channels & $32$             \\
			Number of Recurrent Layers     & $2$              \\
			Convolution Dilations          & $[1,2,4,8,16]$   \\
			Recurrent Decoder Size         & $512$            \\
			Multi-layer Perception Size    & $512$            \\
			Activation                     & ReLU             \\
			\bottomrule
		\end{tabular}
		\label{table:rl-qot-hyperparamters}
	\end{center}
\end{table}

\end{document}